\documentclass[11pt]{article}

\usepackage[preprint]{acl}

\usepackage{times}
\usepackage{latexsym}

\usepackage[T1]{fontenc}

\usepackage[utf8]{inputenc}

\usepackage{microtype}

\usepackage{inconsolata}

\usepackage{hyperref}
\usepackage{url}

\usepackage{graphicx}  
\usepackage{caption}   
\usepackage{float}     
\usepackage{booktabs}   
\usepackage{multirow}   
\usepackage{caption}    
\usepackage{graphicx}   
\usepackage{amsmath}    
\usepackage{array}      

\usepackage[most]{tcolorbox}
\usepackage{xcolor}
\usepackage[table]{xcolor}

\usepackage[T1]{fontenc}
\usepackage{inconsolata} 

\usepackage{tikz}

\newcommand{\circnum}[1]{%
\tikz[baseline=(char.base)]{
\node[
    shape=circle,
    draw,
    fill=blue!20,
    inner sep=2pt,
    font=\small
] (char) {#1};}}

\title{PRISM: Agentic Retrieval with LLMs for Multi-Hop Question Answering}

\author{Md Mahadi Hasan Nahid \\
  University of Alberta\\
  \texttt{mnahid@ualberta.ca} \\\And
  Davood Rafiei \\
  University of Alberta \\
  \texttt{drafiei@ualberta.ca} \\}

\begin{document}
\maketitle

\begin{abstract}
Retrieval plays a central role in multi-hop question answering (QA), where answering complex questions requires gathering multiple pieces of evidence. We propose \textbf{PRISM}, an agentic retrieval framework that leverages large language models (LLMs) in a structured loop to retrieve relevant evidence with high precision and recall. PRISM decomposes retrieval into three specialized agents: a Question Analyzer that breaks complex queries into sub-questions, a Selector that identifies the most relevant context for each sub-question (focusing on precision), and an Adder that brings in any missing evidence (focusing on recall). The iterative interaction between the Selector and Adder produces a compact yet comprehensive evidence set, avoiding both brittle error propagation and noisy context accumulation. It achieves higher retrieval accuracy while filtering out distracting content, enabling downstream QA models to surpass full-context answer accuracy while relying on significantly less irrelevant information. Experiments on four challenging multi-hop QA benchmarks, including HotpotQA, 2WikiMultiHopQA, MuSiQue, and MultiHopRAG, demonstrate that our approach consistently outperforms strong baselines.
\end{abstract}

\section{Introduction}

Retrieval systems lie at the heart of question answering, yet designing them remains highly challenging \citep{asai2024selfrag}. Their effectiveness depends on balancing precision, which ensures that retrieved passages are relevant and free from distracting noise, and recall, which ensures that no essential evidence is missed \citep{trivedi-etal-2023-interleaving}. This trade-off is especially critical in multi-hop reasoning, where evidence is distributed across multiple passages and missing one step can break the reasoning chain while excessive irrelevant context can obscure the signal and degrade performance \citep{lee-etal-2025-shifting}. For example, answering “Which painter who shared a house with Vincent van Gogh was married to a Danish ceramist?” requires first identifying who shared a house with van Gogh (first hop) and then finding whom that person married (second hop). Solving such queries is challenging because a system must retrieve the relevant evidence for each reasoning step while ignoring the many distractors present in a large corpus like Wikipedia \citep{yang-etal-2018-hotpotqa, trivedi-etal-2022-musique, tang2024multihoprag}.

With the advent of large language models (LLMs), these challenges are amplified. LLMs struggle with the \emph{lost-in-the-middle} phenomenon, overlooking crucial evidence buried in long contexts \citep{liu-etal-2024-lost}, and are prone to hallucination when information is incomplete or noisy \citep{laban-etal-2024-summary}. Thus, enhancing retrieval to provide compact, comprehensive, and faithful evidence is essential for reliable reasoning in LLM-based systems.


\begin{figure*}[ht] 
    \centering
    \includegraphics[width=.90\linewidth]{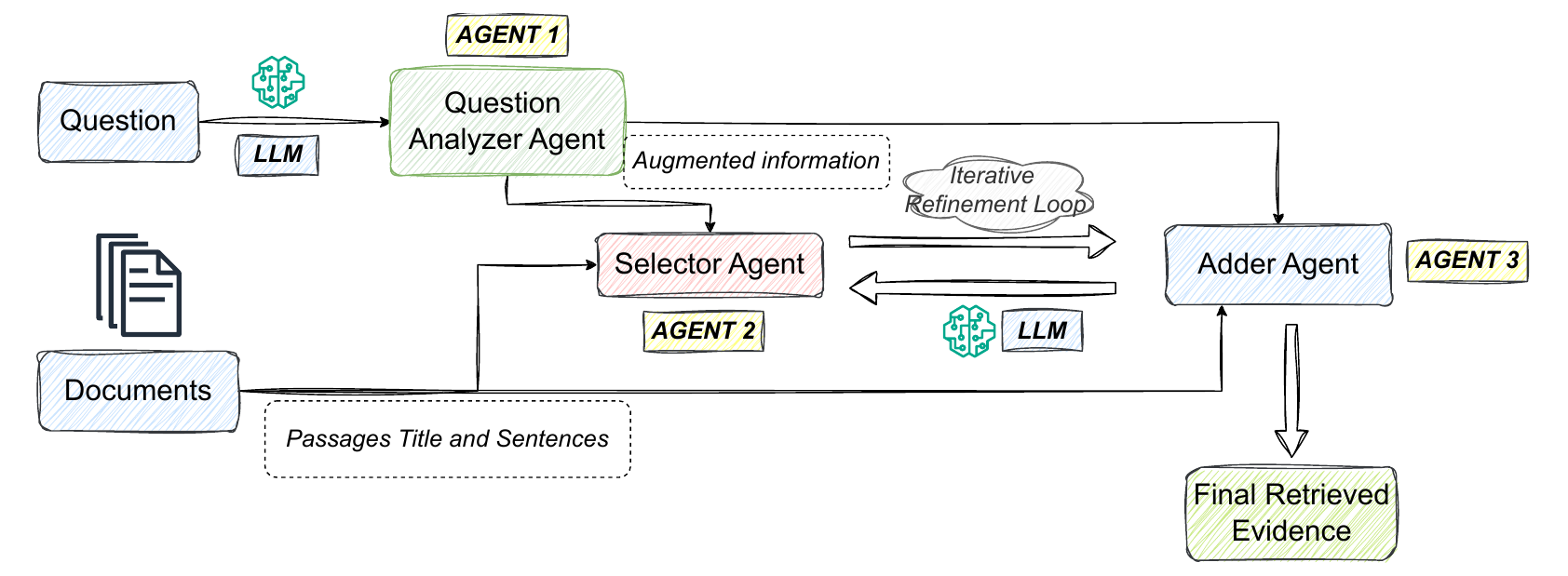}
    \caption{Overview of Agentic Retrieval Framework (PRISM). The complex question is decomposed by the Analyzer into sub-questions, the Selector narrows down relevant evidence for precision, and the Adder expands context for recall. The loop iterates $N$ times to produce a refined evidence set for QA.}
    \label{fig:framework_overview}

\end{figure*}




Early pipeline approaches such as GraphRetriever~\citep{Asai2020Learning} and Multi-hop Dense Retrieval~\citep{xiong2021answering} relied on iterative query rewriting or entity linking to follow reasoning chains~\citep{qi-etal-2019-answering}. While effective in controlled settings, these methods suffer from error propagation, where early mistakes degrade final performance~\citep{sun-etal-2023-chatgpt}. Later frameworks such as ReAct \citep{yao2023react} and IRCoT \citep{trivedi-etal-2023-interleaving} coupled large language models with retrieval, interleaving chain-of-thought \citep{wei2022chain} reasoning and evidence gathering. These approaches improved recall by letting intermediate reasoning guide the search, but often at the cost of large, noisy evidence sets, since precision mechanisms were limited. IRCoT~\citep{trivedi-etal-2023-interleaving} in particular emphasizes recall, leading to distractor-heavy contexts that can obscure reasoning.

Complementary efforts have aimed to improve relevance and coverage through reranking~ \citep{pradeep2023rankzephyr} and Pruning-based methods~\citep{chirkova2025provence}. Likewise, set-wise selection suggests that moving beyond naive top-$k$ retrieval by incorporating reasoning, selection, and refinement is crucial for complex reasoning \citep{lee-etal-2025-shifting}. However, these single-pass strategies are limited in that if essential evidence is absent from the retrieval pool, there is \emph{no mechanism for recovering it} \citep{lee-etal-2025-shifting}. Iterative self-refinement~\citep{asai2024selfrag} attempt to balance these trade-offs, but lack a principled way to separate precision from recall, often generating redundant or hallucinated context. A critical limitation of several baselines~\citep{pradeep2023rankzephyr,sun-etal-2023-chatgpt, yu2024rankrag, ammann2025question} is that they underperform even relative to \emph{full context basline} because their retrieval pipelines fail to preserve the necessary supporting evidence. 


Despite the advances, a key challenge remains: constructing evidence sets that are both \emph{comprehensive and concise}. Standard retrievers and rerankers typically return a top-$k$ list of individually relevant passages \citep{lee-etal-2025-shifting}, but the set as a whole is often redundant or incomplete \citep{pradeep2023rankzephyr, yu2024rankrag}. More broadly, retrieval-augmented generation faces fundamental limitations: (i) single-vector embeddings cannot represent all possible query–document relevance patterns as candidate sets grow combinatorially \citep{weller2026on}; (ii) long-context LLMs systematically overlook mid-context evidence \citep{liu-etal-2024-lost}; and (iii) irrelevant material in long contexts increases hallucination \citep{laban-etal-2024-summary}. Together, these findings highlight that simply scaling embeddings or extending context windows cannot resolve the precision–recall trade-off.


In this work, we introduce \textbf{PRISM} (Precision–Recall Iterative Selection Mechanism), an LLM based agentic retrieval framework in which precision and recall focused agents collaborate through iterative refinement to retrieve high-quality evidence sets. Concretely, PRISM employs three collaborating agents: a \textbf{Question Analyzer} that decomposes the question into sub-questions targeting the required facts; a \textbf{Selector} that filters the candidate pool to eliminate distractors; and an \textbf{Adder} that reconsiders unselected candidates to recover any overlooked evidence. The Selector and Adder iterate, refining the evidence set until it is both compact and complete.

Our contributions are threefold. \circnum{\textbf{1}} We introduce \textbf{PRISM}, an agentic retrieval framework that enables effective control of the precision–recall balance in evidence selection for multi-hop question answering through an iterative loop. \circnum{\textbf{2}} We show that PRISM consistently constructs evidence sets that are both compact and comprehensive, achieving higher precision and recall than strong recent baselines on HotpotQA, 2WikiMultiHopQA, MuSiQue, and MultiHopRAG. \circnum{\textbf{3}} We demonstrate that these high-quality retrieved evidence sets lead to stronger end-to-end QA performance, enabling LLM readers to surpass full-context baselines and recent competitive methods on challenging multi-hop benchmarks where distractors often degrade accuracy.


\section{Methodology}
\label{sec:methodology}


We propose \textbf{PRISM} (Precision–Recall Iterative Selection Mechanism), 
an agentic retrieval framework that decomposes the multi-hop evidence gathering process into a sequence of coordinated steps. PRISM employs three LLM-based agents with distinct roles: the \textbf{Question Analyzer}, which decomposes complex queries into sub-questions; the \textbf{Selector}, which filters candidate evidence to maximize precision; and the \textbf{Adder}, which supplements missing evidence to improve recall. 
\tcbset{
  findingbar/.style={
    colback=green!10,
    colframe=black!70,
    boxrule=1pt,
    left=2pt,
    right=2pt,
    top=2pt,
    bottom=2pt,
    width=\linewidth
  }
}
\vspace{-0.1cm}

\begin{tcolorbox}[findingbar] 
\textbf{Motivating Example.} Consider the question: "Which painter who shared a house with Vincent van Gogh was married to a Danish ceramist?" A naive retriever might return passages about Van Gogh's life or Danish art, many of which are surface-relevant but ultimately distracting. The Question Analyzer decomposes the query into sub-questions such as "Who shared a house with Van Gogh?" and "Who was that person married to?" The Selector retains passages directly answering these sub-questions, such as Paul Gauguin's stay in Arles, while discarding unrelated content. If the bridging passage about Gauguin's marriage is initially missed, the Adder recovers it. After a few iterations, the evidence set contains precisely the passages needed to answer the question correctly.
\end{tcolorbox}
\vspace{-0.1cm}

Each agent is instantiated as a prompted LLM with instructions tailored to its task. Operating in an iterative loop, these agents collectively produce a compact yet comprehensive set of supporting passages. \textbf{Figure~\ref{fig:framework_overview}} illustrates the overall architecture and data flow. We provide an illustrative example in the \textbf{Appendix \ref{app:illustrative_example}}.

\subsection{Question Analyzer Agent}

Multi-hop questions often intertwine multiple entities, relations, and constraints into a single complex query, making it easy for retrievers or LLMs to chase partial matches or irrelevant context. The Question Analyzer mitigates this risk by decomposing the query into a structured set of sub-questions that make the evidence requirements explicit and define a focused search space for later agents \citep{fu-etal-2021-decomposing-complex}. This decomposition reduces the chance of missing critical hops while preventing downstream stages from being overwhelmed with loosely relevant candidates. Concretely, we prompt an LLM with a decomposition template that encourages explicit reasoning, and it returns a list of sub-questions, each targeting a distinct factual unit required to answer the original query.


\subsection{Selector Agent}

The Selector serves as a precision-focused filter that removes distractors introduced during retrieval. Large corpora inevitably yield passages that share surface-level words with the query but are semantically off-topic, and passing such noise directly to a QA model leads to error propagation and wasted context budget. The Selector mitigates this by enforcing a high bar for inclusion: it anchors the reasoning chain with the relevant passages that are strongly aligned with the sub-questions. 

Concretely, the agent is prompted with detailed instructions, the sub-questions, and the (retrieved-)  passages, and tasked with  outputting only those that directly support the query. This filtering stage yields evidence sets with very high precision, ensuring that downstream reasoning operates over reliable, compact inputs and mitigating hallucinations as well as lost-in-the-middle effects. The trade-off, however, is that some relevant passages may be excluded if their connection is less explicit—a limitation addressed by the Adder agent in the next stage.


\subsection{Adder Agent}
While precision is vital, overly strict filtering risks omitting complementary or bridging facts that are crucial for multi-hop reasoning. The Adder is introduced to counterbalance this risk by explicitly prioritizing recall. Its role is to audit what the Selector left behind and add evidence that fill logical gaps.
This two-phase design, ensures that the final evidence set is both compact and complete. The instructions for the Adder agent are similar to the Selector’s, but with a different emphasis. We provide it with the same list of candidates, the sub-questions, and additionally indicate which evidence have already been selected by the Selector. 

In essence, the Adder is a second pass through the candidates, but primed to look for anything important that was left out. One common scenario is when the answer to the sub-question actually requires combining two facts from different passages. The Selector might pick one of them (e.g. a passage that gives part of the answer), not realizing another evidence from different passage is also needed for the full answer. The Adder, seeing what’s selected and what the question is, can identify the complementary piece. In the ideal outcome, after the Adder, the union of selected + added passages equals the complete set of supporting documents needed for answering the question. If the Selector already did a perfect job, the Adder will simply output nothing new. But if something was missing, the Adder ensures recall goes up. This two-phase selection is analogous to having a strict filter followed by a gentle filter: one says “only keep absolutely sure things”, the next says “don’t leave out anything that might be required”.

\subsection{Iteration and Multi-Hop Handling}
Multi-hop questions typically require reasoning over multiple intermediate facts. After decomposition by the Question Analyzer, we provide the original question and all generated sub-questions together to the Selector and Adder agents. This joint formulation allows the Selector$\Leftrightarrow$Adder Cycle to identify the minimal set of evidence that collectively cover the reasoning chain.  

The Selector$\Leftrightarrow$Adder cycle is repeated for several iterations, with the evidence set refined at each step. At the end of the process, evidences from all iterations are merged and de-duplicated, producing a compact set that typically contains only the essential evidence required for the multi-hop question while maintaining high precision and recall. This final set is then passed to a  Answer Generator agent. 

\subsection{Answer Generator Agent}
To demonstrate the overall question answering performance of our retrieval framework we introduce an Answer Generator agent (\textbf{QA Agent}), which produces the answer given the retrieved evidence set using our proposed framework. After the Question Analyzer, Selector, and Adder agents collaboratively construct a compact but comprehensive set of supporting evidence, we provide this context along with the original question to a large language model instructed to generate the final answer. We implement the Answer Generator as a prompted LLM operating in a \textbf{zero-shot} setting, without task-specific fine-tuning. 
This design choice allows us to directly assess how improvements in retrieval quality translate into downstream QA performance, while avoiding additional supervision or domain adaptation.

\section{Experimental Setup}

\subsection{Datasets}
We evaluate our framework on four standard multi-hop QA benchmarks in the open-domain setting. \textbf{HotpotQA} \citep{yang-etal-2018-hotpotqa} requires reasoning over multiple Wikipedia paragraphs, and we use its open-domain variant with full-corpus retrieval. \textbf{2WikiMultiHopQA} \citep{ho-etal-2020-constructing} connects pairs of Wikipedia pages via shared entities; following prior work, we use the structured text component of the development set. \textbf{MuSiQue} \citep{trivedi-etal-2022-musique} contains 2–4 hop questions designed to prevent reasoning shortcuts. \textbf{MultiHop-RAG} \citep{tang2024multihoprag} introduces multi-hop queries over a news-article corpus and is used to analyze retrieval precision and recall in a RAG setting. \footnote{For evaluation, we sample 500 instances from each dataset, constrained by computational budget, while ensuring that the subset remains representative of the overall distribution. }


\subsection{Evaluation Metrics and Baselines}

\paragraph{Evaluation Metrics.}
We assess retrieval using standard supporting fact metrics, treating each passage as the retrieval unit. A passage is correct if it contains at least one gold supporting sentence (HotpotQA, 2WikiMultihopQA) or matches a labeled supporting paragraph (MuSiQue, MultiHopRAG). We report \textbf{Precision} (fraction of retrieved passages that are gold) and \textbf{Recall} (fraction of gold passages retrieved), and also track 
the number of retrieved passages, as compact evidence sets improve efficiency and reduce noise. For HotpotQA and 2WikiMultihopQA, we additionally report \textbf{fact-wise precision and recall} at the sentence level. End-to-end QA is evaluated with \textbf{Exact Match (EM)} and token-level \textbf{F1}. We further measure performance under three conditions: (i) \textit{full context}, where the QA model sees all passages including distractors, (ii) \textit{retrieved evidence only}, using the filtered set from our framework, and (iii) \textit{gold evidence only}, where the QA agent receives only the gold supporting facts. 




\paragraph{Baselines.}
We compare our Agentic Retrieval framework against strong baselines spanning iterative retrieval, reranking, and supervised methods. \textbf{IRCoT} \citep{trivedi-etal-2023-interleaving} interleaves chain-of-thought reasoning with iterative retrieval, while \textbf{SetR} \citep{lee-etal-2025-shifting} performs set-wise LLM-based reranking. We further compare against unsupervised and prompt-based retrievers, including \textbf{BM25} \citep{robertson2009probabilistic}, \textbf{DSP} \citep{khattab2022demonstrate}, \textbf{DecomP} \citep{khot2023decomposed}, \textbf{RankZephyr} \citep{pradeep2023rankzephyr}, and \textbf{RankGPT} \citep{sun-etal-2023-chatgpt}, as well as recent supervised and semi-supervised approaches: \textbf{RECOMP} \citep{xu2024recomp}, \textbf{CoRAG} \citep{wang2025chainofretrieval}, \textbf{R1-Researcher} \citep{R1-searcher}, \textbf{O$^2$-Searcher} \citep{mei2025o2searcher}, \textbf{ARM} \citep{chen-etal-2025-retrieve}, \textbf{QD-RAG} \citep{ammann2025question}, and \textbf{RankRAG} \citep{yu2024rankrag}.

\subsection{Implementation Details}
\label{sec:llms}
PRISM is a fully LLM-based, training-free retrieval framework that operates over an initial candidate pool and focuses on evidence selection and refinement. In our experiments, this initial pool is the full candidate set provided by each benchmark (e.g., 10 candidates per question in HotpotQA, 20 in MuSiQue). PRISM targets the second stage of retrieval, where the goal is to construct a compact yet comprehensive evidence set by balancing precision and recall from the candidate pool.

We implement all agents as prompted large language models. Unless stated otherwise, all experiments use \texttt{GPT-4o} \citep{OpenAI_gpt4o_system_card_2024}. We additionally report results using \texttt{Gemini-2.5-Flash-Lite} \citep{google_gemini2.5_report_2025} and \texttt{DeepSeek-Chat} \citep{deepseek_ai_deepseek-v3_2024}, all of which are state-of-the-art large-context models capable of processing the full candidate passage set during evaluation. Prompting is performed in a zero-shot setting, without task-specific fine-tuning. We enforce consistent structured outputs (lists of indices/titles) to ensure reliable parsing. The Selector$\Leftrightarrow$Adder cycle is repeated at most $N=3$ iterations, keeping the number of LLM calls tractable. The rationale for selecting $N=3$ is discussed in Section~\ref{sec:results} (Figure \ref{fig:iter_N}). The detailed propmts are provided in the Appendix~\ref{app:prompts}. 
The implementation of our framework is publicly available at \url{https://github.com/mahadi-nahid/PRISM}.



\section{Results and Discussion}
\label{sec:results}

\subsection{Main Result}

\paragraph{Retrieval Performance.}

\begin{figure}[ht]
    \centering

    \includegraphics[width=0.90\linewidth]{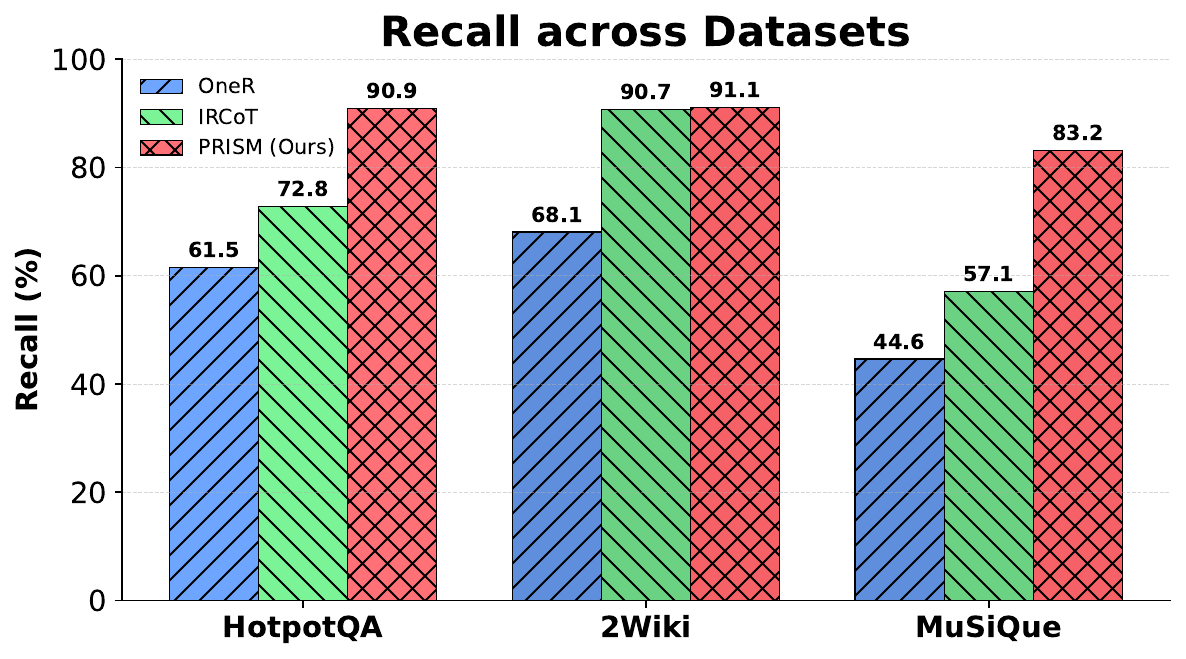}
 
    \caption{ Passage recall across HotpotQA, 2WikiMultiHopQA, and MuSiQue. Our agentic retrieval framework (PRISM) consistently outperforms OneR and IRCoT \citep{trivedi-etal-2023-interleaving}, with especially large gains on the challenging MuSiQue benchmark.}
    \label{fig:recall}

\end{figure}

Figure~\ref{fig:recall} reports recall performance across HotpotQA, 2WikiMultihopQA, and MuSiQue dataset. Our agentic retrieval framework achieves consistently higher recall than both OneR (one pass retriever), which utilize BM25 \citep{robertson2009probabilistic} to retrieves $K$ paragraphs using the original question as a single query, and IRCoT, which guides retrieval through Chain-of-Thought reasoning, across all benchmarks  \citep{trivedi-etal-2023-interleaving}. On HotpotQA, our method attains \textbf{90.9\%} recall compared to 61.5\% for OneR and 72.8\% for IRCoT. On 2WikiMultihopQA, we reach \textbf{91.1\%} recall, improving over OneR (68.1\%) and slightly exceeding IRCoT (90.7\%).  The largest margin appears on MuSiQue, where our approach achieves \textbf{83.2\%} recall versus 44.6\% and 57.1\% for OneR and IRCoT, respectively. 
These gains validate the importance of explicitly balancing precision and recall through our Selector$\Leftrightarrow$Adder loop: the Selector removes distractors while the Adder recovers missed but necessary passages, ensuring a comprehensive evidence set. 


\begin{table}[ht]
\centering
\small

\begin{tabular}{l|c|c}
\toprule
\textbf{Method} & \textbf{Prec.} & \textbf{Recall} \\
\midrule
BM25 (\citeyear{robertson2009probabilistic}) & 11.09 & 24.13 \\
{bge-large-en-v1.5} (\citeyear{xiao2023cpack}) & 16.12 & 32.32 \\
RankGPT (GPT-4o) (\citeyear{sun-etal-2023-chatgpt}) & 17.99 & 36.01 \\
SETR-CoT \& IRI (\citeyear{lee-etal-2025-shifting}) & \underline{22.68} & \underline{36.69} \\
\midrule
\rowcolor{gray!10}
\textbf{PRISM (Ours)} & \textbf{{28.18}} & \textbf{{42.22}} \\
\bottomrule
\end{tabular}

\caption{Retrieval performance on the MultiHopRAG dataset. Our agentic retrieval framework substantially outperforms all baselines, achieving the highest recall while maintaining competitive precision.}

\label{tab:retrieval_multihoprag}
\end{table}


\begin{table*}[ht]
\centering
\small
\setlength{\tabcolsep}{4pt} 

\resizebox{0.90\linewidth}{!}{
\begin{tabular}{l|cc|cc|cc|cc}
\toprule
\multirow{2}{*}{\textbf{Method}} & \multicolumn{2}{c|}{\textbf{HotpotQA}} & \multicolumn{2}{c|}{\textbf{2Wiki}} & \multicolumn{2}{c|}{\textbf{MuSiQue}} & \multicolumn{1}{c}{\textbf{MHRAG}} \\
\cmidrule(lr){2-3} \cmidrule(lr){4-5} \cmidrule(lr){6-7} \cmidrule(lr){8-8}
 & EM & F1 & EM & F1 & EM & F1 & ACC \\
\midrule
\rowcolor{red!10}
Full Context (without retrieval) & 44.18 & 58.28 & 43.20 & 52.11 & 19.77 & 29.42 & 44.37 \\
\midrule
\rowcolor{green!10}
Oracle Context & 64.80 & 77.83 & 61.40 & 71.10 & 38.78 & 50.89 & 57.04 \\
\midrule
DSP \citep{khattab2022demonstrate} &  \underline{51.4}  & \underline{62.9}  & - & - & - & - & - \\
DecomP \citep{khot2023decomposed} &  -  & 53.5  & - & \textbf{70.8} & - & 30.9 & - \\
ZeroR \citep{trivedi-etal-2023-interleaving} &  -  & 41.0 & - & 38.5 & - & 19.0 & - \\
OneR \citep{trivedi-etal-2023-interleaving} &  -   & 50.7  & - & 46.4 & - & 20.4 & - \\
IRCoT QA \citep{trivedi-etal-2023-interleaving} & {49.3}  & {60.7}  & \textbf{57.7} & \underline{68.0}  & \underline{26.5} & \underline{36.5}  & - \\

RankZephyr \citep{pradeep2023rankzephyr} & 34.69 & 35.04 & 33.87 & 27.83 & 8.61 & 12.79 & 43.90 \\ 
RankZephyr + CoT  \citep{pradeep2023rankzephyr} & 33.99 & 34.38 & 33.66 & 27.85 & 9.43 & 13.27 & 43.60 \\ 

RankGPT \citep{sun-etal-2023-chatgpt, lee-etal-2025-shifting} & 34.61 & 35.26 & 34.77 & 28.18 & 9.52 & 13.51 & 45.26 \\
SETR-CoT \& IRI \citep{lee-etal-2025-shifting}  & 39.16 & 40.49 & 35.68 & 31.09 & 12.33 & 16.91 & 47.14 \\
\midrule
\rowcolor{gray!10}
\textbf{PRISM + QA Agent (Ours)} & \textbf{54.20} & \textbf{66.96} & \underline{48.60} & {56.97} & \textbf{31.17} & \textbf{41.78} & \textbf{49.16} \\
\bottomrule
\end{tabular}
}

\caption{End-to-end QA performance on HotpotQA, 2Wiki, MuSiQue and MultiHopRAG. Metrics are Exact Match (EM) and token-level F1. Our agentic retrieval framework (PRISM) achieves state-of-the-art accuracy on HotpotQA, MuSiQue and MultiHopRAG, while remaining competitive on 2Wiki.}

\label{tab:qa}
\end{table*}


We also assessed our proposed framework’s retrieval performance using the MultiHopRAG dataset  \citep{tang2024multihoprag} following \citet{lee-etal-2025-shifting}. Table~\ref{tab:retrieval_multihoprag} summarizes retrieval results on the MultiHopRAG dataset. Our agentic retrieval framework achieves a precision of 28.18 and a recall of 42.22, which represents a large improvement over all baselines. Classical BM25 \citep{robertson2009probabilistic} performs poorly with only 11.09 precision and 24.13 recall, while dense retrievers such as \texttt{bge-large-en-v1.5} \citep{xiao2023cpack} and RankGPT \citep{sun-etal-2023-chatgpt} improve recall to around 32--36 but still fall significantly short of our method. SETR-CoT \& IRI \citep{lee-etal-2025-shifting} narrows the gap with 22.68 precision and 36.69 recall, yet our framework surpasses it by about 6 points in both recall and precision. These gains highlight that the combination of precision-oriented selection and recall-oriented addition in our agentic loop is also effective for the MultiHopRAG dataset, which demands accurate recovery of multiple supporting evidence across hops.

\paragraph{Impact on QA Accuracy.}

Table~\ref{tab:qa} shows that our agentic retrieval framework consistently improves end-to-end QA. On \textbf{HotpotQA} and \textbf{MuSiQue}, it surpasses recent methods such as IRCoT \citep{trivedi-etal-2023-interleaving} and SetR  \citep{lee-etal-2025-shifting} by clear margins (e.g., +5 EM / +6 F1 on HotpotQA), confirming that filtering distractors while recovering missing facts strengthens multi-hop reasoning. On \textbf{MultiHopRAG}, our method achieves the best accuracy (49.16), highlighting the robustness of the Selector$\Leftrightarrow$Adder balance for distributed evidence. While \textbf{2WikiMultihopQA} favors recall-heavy approaches like IRCoT, our system remains competitive and outperforms other baselines \footnote{Our QA agent operates in a zero-shot setting; no few-shot demonstrations were used in our experiments like IRCoT-QA. }. Our framework also surpasses full-context baselines across all datasets by a significant margin and narrows the gap to the oracle-context setting, where the LLM is provided only with the gold supporting evidence required to answer the question. Overall, these results validate that retrieval quality is a decisive factor for QA performance, with precision--recall balancing yielding compact, effective evidence sets. 

We also observe a critical limitation in several baselines: they often underperform compared to the Full Context setting (w/o retrieval) because their retrieval pipelines fail to preserve the necessary supporting evidence in the final context. As a result, downstream QA performance drops even when the required evidence is available in the candidate pool. PRISM mitigates this issue by operating over the full candidate pool provided by each dataset and refining it through iterative selection, which improves evidence completeness while reducing noise.

\paragraph{Performance with Alternative LLMs.}

To assess the robustness of our framework across different large language models, we evaluated retrieval and QA performance using \texttt{Gemini-2.5-Flash-Lite} and \texttt{DeepSeek}, in addition to \texttt{GPT-4o}. 

Table~\ref{tab:llm_results} shows that our framework maintains strong retrieval and QA performance across different LLM backends. While absolute scores vary, the precision--recall balancing of the Selector--Adder loop consistently yields high recall and competitive accuracy. These results confirm that our agentic retrieval framework generalizes across LLMs, and improvements in base model reasoning naturally translate into retrieval and downstream QA gains.



\begin{table*}[ht]
\centering
\footnotesize
\setlength{\tabcolsep}{4pt} 

\resizebox{0.90\linewidth}{!}{
\begin{tabular}{l|c|c|c|c|c|c}
\toprule
\multirow{2}{*}{\textbf{LLM}}
& \multicolumn{2}{c|}{\textbf{HotpotQA}} 
& \multicolumn{2}{c|}{\textbf{2Wiki}} 
& \multicolumn{2}{c}{\textbf{MuSiQue}} \\
\cmidrule(lr){2-3} \cmidrule(lr){4-5} \cmidrule(lr){6-7}
 & P/R & EM/F1 & P/R & EM/F1 & P/R & EM/F1 \\
\midrule
\texttt{GPT-4o} & \underline{83.02}/90.90 & 54.20/66.96 & 90.97/91.07  & \textbf{48.60}/\textbf{56.97} & 47.46/83.17 & 31.17/41.78 \\
\midrule
\texttt{Gemini-2.5-Flash-Lite} & \textbf{87.24}/\underline{93.46} & \underline{56.93}/\underline{70.37} & \textbf{93.73}/\underline{95.19}  & \underline{47.65}/\underline{55.58}  & \underline{63.06}/\underline{83.74}  & \textbf{36.64}/\textbf{44.52} \\
\midrule
\texttt{DeepSeek-chat}            & 79.46/\textbf{95.88}    & \textbf{57.60}/\textbf{70.62}    & \underline{93.01}/\textbf{97.38}  & 43.39/54.75  & \textbf{67.11}/\textbf{89.38}  & \underline{31.93}/\underline{42.45}  \\
\bottomrule
\end{tabular}
}

\caption{Retrieval (P/R) and QA (EM/F1) performance of our framework with different LLM backends on HotpotQA, 2WikiMultiHopQA, and MuSiQue. Results show that our framework consistently maintains high recall and competitive QA accuracy.}

\label{tab:llm_results}
\end{table*}

\begin{table}[ht]
\centering
\small
\setlength{\tabcolsep}{4pt} 

\resizebox{0.95\linewidth}{!}{

\begin{tabular}{l|cc|cc|cc}
\toprule
\multirow{2}{*}{\textbf{Method}} & \multicolumn{2}{c|}{\textbf{HotpotQA}} & \multicolumn{2}{c|}{\textbf{2Wiki}} & \multicolumn{2}{c}{\textbf{MuSiQue}} \\
\cmidrule(lr){2-3} \cmidrule(lr){4-5} \cmidrule(lr){6-7}
 & EM & F1 & EM & F1 & EM & F1  \\
\midrule

RECOMP (\citeyear{xu2024recomp}) &  28.20 & 37.91 & - & - & - & -  \\
CoRAG (\citeyear{wang2025chainofretrieval}) &  \textbf{56.30} & \textbf{69.80}  & \textbf{72.50} & \textbf{77.30} & \underline{30.90} & \textbf{42.40}  \\

R1-Researcher (\citeyear{R1-searcher}) &  -  & 34.20 & - & 34.40 & - & 17.20  \\
O$^2$-Searcher (\citeyear{mei2025o2searcher}) &  -   & 38.80   & - & 37.40 & - & 16.00  \\

ARM (\citeyear{chen-etal-2025-retrieve}) &  -  & - & - & 71.70  & - & -  \\
QD-RAG (\citeyear{ammann2025question}) &  28.10   & 35.00   & - & - & - & -  \\
RankRAG (\citeyear{yu2024rankrag}) &  35.30 & 46.70   & 31.40 & 36.90 & - & - \\
\midrule
\rowcolor{gray!10}
\textbf{PRISM+QA} & \underline{54.20} & \underline{66.96} & \underline{48.60} & \underline{56.97} & \textbf{31.17} & \underline{41.78}  \\
\bottomrule
\end{tabular}
}

\caption{End-to-end QA performance on HotpotQA, 2Wiki, MuSiQue dataset on more recent fine-tuned and compression based methods. Metrics are Exact Match (EM) and token-level F1. }

\label{tab:additional_qa_results}
\end{table}
\paragraph{Comparison with Finetuned and Compression Based Methods.}

Table \ref{tab:additional_qa_results} summarizes the end-to-end QA performance of PRISM against a range of contemporary fine-tuned and comparison based models. From the table we can see that on HotpotQA, PRISM's 66.96 F1 significantly outperforms context compression based method RECOMP and general ranking methods RankRAG. Notably, PRISM achieves performance highly competitive with the costly fine-tuned model CoRAG. On the challenging MuSiQue benchmark, PRISM achieves strong F1 score (41.78) in this comparison, validating that our iterative Selector-Adder loop effectively constructs the comprehensive reasoning chains needed for deep multi-hop QA, outperforming specialized RL-trained search agents \citep{mei2025o2searcher}. While fine-tuned models excel on 2Wiki, the strong performance of PRISM (56.97 F1) reinforces its core value proposition: superior performance-to-cost trade-off by intelligently leveraging LLM reasoning in a novel agentic structure, without requiring expensive fine-tuning.


\begin{figure}[ht]
    \centering
    \resizebox{.90\linewidth}{!}{
    \includegraphics[width=0.95\linewidth]{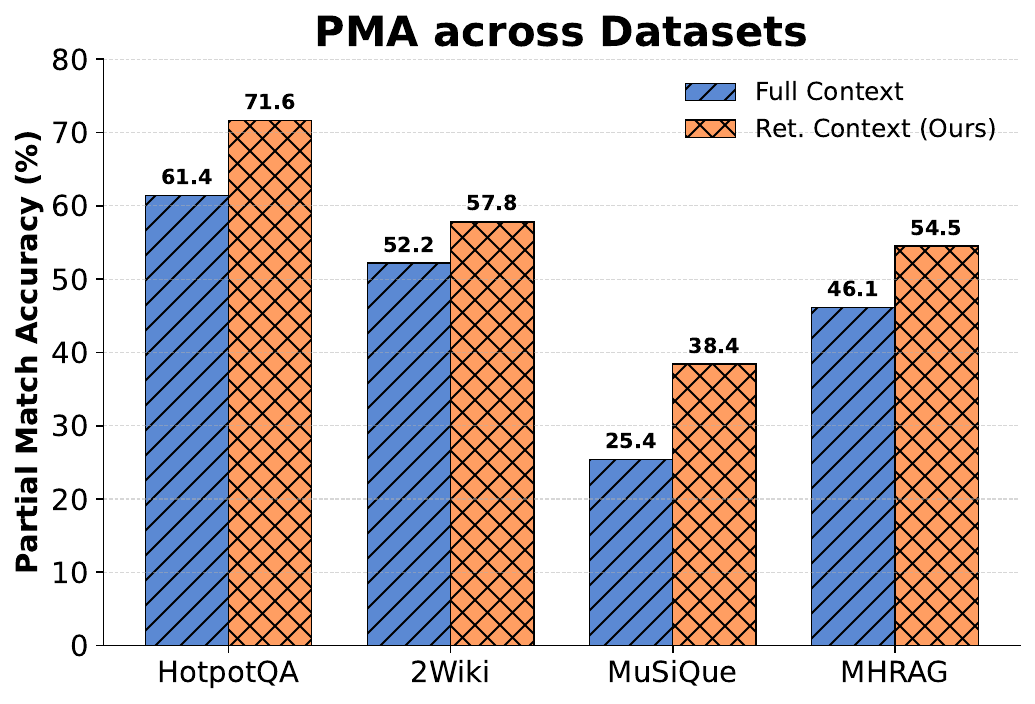}
    }
    \caption{ Partial Match Accuracy (PMA) of our QA agent across four datasets. Our retrieval-based approach improves robustness on all datasets by reducing distractors and focusing the generator on relevant content.}
    \label{fig:pma}
\end{figure}

\paragraph{Partial Match Accuracy.} 

Figure~\ref{fig:pma} shows that our framework achieves consistent gains in Partial Match Accuracy (PMA) across all four datasets. Compared to the full-context baseline, our retrieved evidence improves PMA by 10 points on HotpotQA and more than 10 points on MuSiQue, with consistent gains also observed on 2WikiMultiHopQA and MultiHopRAG.   
These results highlight that filtering out distractors not only improves exact matching but also increases the likelihood of partial match demonstrating the robustness of our retrieval-based approach.


\begin{table*}[ht]
\centering
\small
\setlength{\tabcolsep}{8pt} 

\begin{tabular}{l|c|c|c|c|c|c}
\toprule
\multirow{2}{*}{\textbf{Variant}} 
& \multicolumn{2}{c|}{\textbf{HotpotQA}} 
& \multicolumn{2}{c|}{\textbf{2Wiki}} 
& \multicolumn{2}{c}{\textbf{MuSiQue}} \\
\cmidrule(lr){2-3} \cmidrule(lr){4-5} \cmidrule(lr){6-7}
 & P/R & \#Pssg & P/R & \#Pssg & P/R & \#Pssg \\
\midrule
Full Model & 83.0/\textbf{90.9} & 2.67 & 83.2/\textbf{91.1} & 2.74 & 47.7/\textbf{83.2} & 6.15 \\
\midrule
\texttt{w/o Q. Analyzer} & 78.9/\textbf{86.8} & 2.88 & 90.6/\textbf{85.8} & 2.68 & 36.2/\textbf{68.8} & 7.68 \\
\midrule
\texttt{w/o Adder (Selector Only)} & 86.9/\textbf{79.7} & 2.63 & 92.1/\textbf{80.5} & 2.71 & 77.4/\textbf{69.3} & 2.82 \\
\midrule
\texttt{w/o Adder (no-op)}	&  87.9/\textbf{80.0} & 2.62 & 	92.3/\textbf{80.5} & 2.72	&  78.2/\textbf{67.7} & 2.84 \\
\bottomrule
\end{tabular}
\caption{Ablation study on HotpotQA, 2WikiMultihopQA, and MuSiQue showing that both the Question Analyzer and Selector$\Leftrightarrow$Adder loop are crucial, as removing either significantly reduces recall and weakens evidence completeness.}
\label{tab:ablation}
\end{table*}

\begin{figure}[ht]
    \centering
    \resizebox{.90\linewidth}{!}{
    \includegraphics[width=1\linewidth]{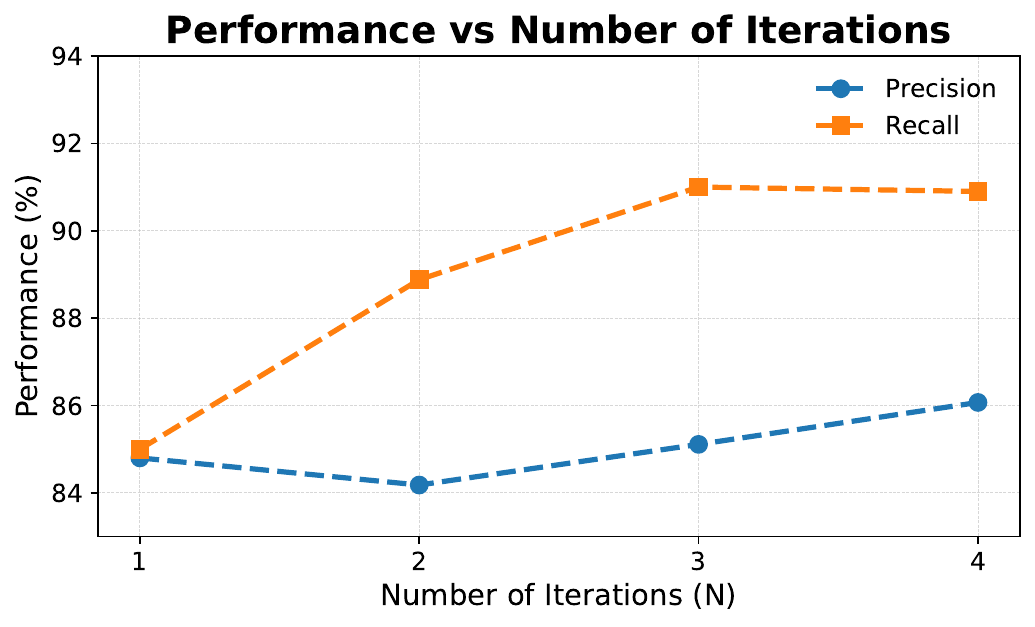}
    }
    \caption{Experiment on Variying Number of Selector$\Leftrightarrow$Adder Iteration on HotpotQA dataset on 100 samples.}
    \label{fig:iter_N}

    \vspace{-0.6em}
\end{figure}

\paragraph{Experiment on Varying Number of Iteration.}

We conducted a sensitivity analysis to determine the optimal number of Selector$\Leftrightarrow$Adder iterations, N, before conducting our experiments. As summarized in Figure~\ref{fig:iter_N} on the HotPotQA dataset, performance generally improves with an increasing number of iterations. However, we observed that the improvement becomes marginal after N=4. We noted that the outputs of the Selector and Adder agents converged after the third iteration, yielding almost identical responses in iterations 3 and 4. Based on this trade-off between performance and computational cost, we selected N=3 for our final experiments.

\paragraph{Ablation Study.} 
Table~\ref{tab:ablation} shows that each component of our framework plays a critical role in achieving high recall. The full model consistently achieves the highest recall across all datasets. Removing the Question Analyzer leads to a substantial drop, particularly on MuSiQue (83.2 $\rightarrow$ 68.8), highlighting the importance of question decomposition for complex multi-hop reasoning.

To better isolate the contribution of the Adder, we include two variants: (i) removing the Adder entirely (Selector-only) and (ii) a no-op Adder. Both variants show a significant reduction in recall (e.g., 90.9 $\rightarrow$ 79.7 and 80.0 on HotpotQA), confirming that the Adder is essential for recovering false negatives. The similar performance between the Selector-only and no-op variants further indicates that the gains are not due to additional computation, but stem from the Adder’s explicit recovery mechanism.

Overall, these results demonstrate that the Question Analyzer and the Selector$\Leftrightarrow$Adder iterative loop are complementary, and together enable PRISM to construct a more complete and balanced evidence set.

\paragraph{Additional Results and Analysis.} Additional results and detailed error analysis are discussed in \textbf{Appendix \ref{app:additional_results_error_analysis}}. 

\subsection{Discussion}

Our results demonstrate that retrieval quality is one of the key bottlenecks in multi-hop QA and that explicitly balancing precision and recall is crucial for robust performance. By separating precision-oriented selection (Selector) from recall-oriented addition (Adder), our framework produces compact yet comprehensive evidence sets that improve answer accuracy across diverse datasets. On HotpotQA and MuSiQue, this balance enables clear gains over prior state-of-the-art methods, while on MultiHopRAG it proves particularly effective at recovering distributed evidence across documents. Even in 2WikiMultiHopQA, where recall-heavy approaches such as IRCoT remain slightly stronger, our framework remains competitive and consistently outperforms other alternatives.

\begin{tcolorbox}[findingbar] 
\circnum{\textbf{1}} {Balancing precision and recall is critical for multi-hop QA.}
\end{tcolorbox}

Beyond dataset-specific results, two broader trends emerge. First, removing distractors helps the QA model focus on relevant reasoning chains, yielding higher exact match and partial match accuracy compared to full-context baselines. Second, the framework generalizes across different LLM backends, with retrieval improvements translating naturally into QA gains regardless of the base model. This confirms that our agentic loop is not tied to a specific architecture, but instead leverages the underlying LLM’s reasoning ability to optimize evidence selection.

\begin{tcolorbox}[findingbar] 
\circnum{\textbf{2}} {Cleaner evidence leads to stronger reasoning and model-agnostic gains.}
\end{tcolorbox}

These findings highlight that retrieval should not be treated as a static preprocessing step, but rather as an active, agent-driven process that collaborates with the QA model. By iteratively refining evidence through complementary precision and recall objectives, our approach provides a principled path toward building reliable, reasoning-centric retrieval systems for multi-hop QA.

\begin{tcolorbox}[findingbar] 
\circnum{\textbf{3}} {Retrieval should be treated as an active, agent-driven process.}
\end{tcolorbox}

In addition to these empirical insights, several broader strengths stand out. The modular multi-agent structure provides interpretability and control over the retrieval process. Another key strength is the framework’s robust generalization across models and datasets: despite differences in absolute performance across GPT-4o, Gemini, and other LLMs, our method consistently maintains high recall and competitive QA accuracy, demonstrating strong model-agnostic adaptability. Moreover, the design is conceptually extensible, naturally accommodating adaptive iteration strategies and domain-specific agents, positioning agentic retrieval as a foundation for next-generation retrieval-augmented reasoning systems.

\begin{tcolorbox}[findingbar] 
\circnum{\textbf{4}} {Modularity enables interpretability and extensibility.}
\end{tcolorbox}

\section{Related Work}
\label{rel-work-main}

\paragraph{Multi-hop QA and Retrieval.} Benchmarks such as HotpotQA \citep{yang-etal-2018-hotpotqa}, 2WikiMultiHopQA \citep{ho-etal-2020-constructing}, and MuSiQue \citep{trivedi-etal-2022-musique} have driven research in multi-hop reasoning over large corpora. Early methods relied on iterative retrieval via entity linking or path traversal (e.g., Graph Retriever \citep{Asai2020Learning}, PathRetriever, MDR \citep{xiong2021answering}), but these pipelines are brittle due to error propagation across hops. Prompt-based frameworks such as DSP \citep{khattab2022demonstrate} and Decomposed Prompting \citep{khot2023decomposed} demonstrated that integrating reasoning with retrieval and decomposing complex questions improves performance. While question decomposition has a long history \citep{talmor-berant-2018-web, min-etal-2019-multi, fu-etal-2021-decomposing-complex, ammann2025question}, prior approaches typically issue independent searches and aggregate noisy evidence without explicit verification.

\paragraph{Passage Selection and Context Optimization.} Recent work emphasizes compact evidence selection. Set-wise and listwise reranking methods such as SetR \citep{lee-etal-2025-shifting} and RankZephyr \citep{pradeep2023rankzephyr} improve coverage over independent scoring, while pruning and refinement approaches (e.g., Provence \citep{chirkova2025provence}, Self-RAG \citep{asai2024selfrag}) reduce spurious context. RankRAG \citep{yu2024rankrag} jointly performs ranking and generation via supervised fine-tuning but incurs substantial training cost. In contrast, our method explicitly decouples precision and recall into separate agents within an iterative loop, enabling controllable evidence expansion without fine-tuning.

\paragraph{Agentic and Finetuned Retrieval.} RL- and SFT-based retrievers such as CoRAG \citep{wang2025chainofretrieval}, R1-Searcher \citep{R1-searcher}, and O$^2$-Searcher \citep{mei2025o2searcher} embed retrieval policies into model weights, achieving strong results at high training cost and limited portability. Other approaches optimize context post-retrieval via compression (e.g., RECOMP \citep{xu2024recomp}, EXIT \citep{hwang-etal-2025-exit}) or structural indexing (e.g., RAPTOR \citep{sarthi2024raptor}). Our framework instead operates purely at inference time, leveraging off-the-shelf LLMs to iteratively balance precision and recall over flat indices for multi-hop QA.

We provide an extended review of related work, including a more detailed comparison with existing retrieval and agentic reasoning approaches, in Appendix~\ref{rel-work-extended}.

\section{Conclusion}

We presented \textbf{PRISM}, an agentic retrieval framework for multi-hop question answering that treats retrieval as an iterative, reasoning-driven process. By decomposing questions and collaborating through a precision-oriented {Selector} and a recall-oriented {Adder} agent, PRISM constructs evidence sets that are both compact and comprehensive, without requiring task-specific fine-tuning. Across four challenging benchmarks, PRISM consistently improves retrieval quality and downstream QA performance, surpassing recent baselines while using significantly less irrelevant context. The framework generalizes across LLM backends and operates purely at inference time, offering a favorable performance--cost trade-off. We believe this work provides a practical foundation for future research on reasoning-centric retrieval in complex, knowledge-intensive tasks.

\section*{Limitations}

Despite its effectiveness, our approach has several limitations that suggest directions for future work. First, the multi-agent architecture incurs higher computational cost than single-pass retrieval pipelines. While this overhead is partially mitigated by compact evidence sets and a bounded number of iterations, improving efficiency remains important for scaling to larger corpora or latency-sensitive settings. 

Second, although the Selector–Adder loop recovers most relevant evidence, it may occasionally miss subtle reasoning chains or introduce redundancy when passages are only weakly connected. Addressing this limitation may require more adaptive iteration control, uncertainty-aware selection mechanisms, or tighter coupling between retrieval decisions and explicit reasoning signals.

Finally, our evaluation is limited to standard multi-hop QA benchmarks. Applying the framework to specialized domains such as scientific, biomedical, or legal corpora may require domain-specific adaptations, including tailored retrieval strategies or specialized prompting. In scenarios where the candidate pool is very large (e.g., thousands of passages), directly applying PRISM may be challenging due to LLM context limitations. Extending PRISM to open-domain retrieval, for example by integrating an external retriever to generate a manageable candidate pool, remains an important direction for future work.

\section*{Ethical Considerations}

This work studies retrieval methods for multi-hop question answering using publicly available benchmarks and corpora, which do not contain sensitive personal data. The proposed framework does not involve data collection, annotation, or model training, and therefore poses minimal privacy risk.

As with other retrieval-augmented systems, the approach may reflect biases present in the underlying corpora or language models. In addition, the system is evaluated only on benchmark datasets and should not be deployed in high-stakes domains without domain-specific validation and appropriate human oversight.


\section*{Acknowledgements}
We sincerely thank the reviewers for their thoughtful feedback, insightful suggestions, and encouraging remarks, which helped improve the quality of this work. This research was supported in part by the Natural Sciences and Engineering Research Council of Canada (NSERC). We also acknowledge the computational resources provided by the Digital Research Alliance of Canada. Md Mahadi Hasan Nahid was supported by the Alberta Innovates Graduate Student Scholarship.


\bibliography{custom}

\begin{thebibliography}{41}
\providecommand{\natexlab}[1]{#1}

\bibitem[{Ammann et~al.(2025)Ammann, Golde, and Akbik}]{ammann2025question}
Paul J.~L. Ammann, Jonas Golde, and Alan Akbik. 2025.
\newblock \href {https://openreview.net/forum?id=ZFxeCvtAqg} {Question decomposition for retrieval-augmented generation}.
\newblock In \emph{ACL 2025 Student Research Workshop}.

\bibitem[{Asai et~al.(2020)Asai, Hashimoto, Hajishirzi, Socher, and Xiong}]{Asai2020Learning}
Akari Asai, Kazuma Hashimoto, Hannaneh Hajishirzi, Richard Socher, and Caiming Xiong. 2020.
\newblock \href {https://openreview.net/forum?id=SJgVHkrYDH} {Learning to retrieve reasoning paths over wikipedia graph for question answering}.
\newblock In \emph{International Conference on Learning Representations}.

\bibitem[{Asai et~al.(2024)Asai, Wu, Wang, Sil, and Hajishirzi}]{asai2024selfrag}
Akari Asai, Zeqiu Wu, Yizhong Wang, Avirup Sil, and Hannaneh Hajishirzi. 2024.
\newblock \href {https://openreview.net/forum?id=hSyW5go0v8} {Self-{RAG}: Learning to retrieve, generate, and critique through self-reflection}.
\newblock In \emph{The Twelfth International Conference on Learning Representations}.

\bibitem[{Chan et~al.(2024)Chan, Xu, Yuan, Luo, Xue, Guo, and Fu}]{chan2024rqrag}
Chi-Min Chan, Chunpu Xu, Ruibin Yuan, Hongyin Luo, Wei Xue, Yike Guo, and Jie Fu. 2024.
\newblock \href {https://openreview.net/forum?id=tzE7VqsaJ4} {{RQ}-{RAG}: Learning to refine queries for retrieval augmented generation}.
\newblock In \emph{First Conference on Language Modeling}.

\bibitem[{Chen et~al.(2025)Chen, Zhang, Cafarella, and Roth}]{chen-etal-2025-retrieve}
Peter~Baile Chen, Yi~Zhang, Mike Cafarella, and Dan Roth. 2025.
\newblock \href {https://doi.org/10.18653/v1/2025.acl-long.1463} {Can we retrieve everything all at once? {ARM}: An alignment-oriented {LLM}-based retrieval method}.
\newblock In \emph{Proceedings of the 63rd Annual Meeting of the Association for Computational Linguistics (Volume 1: Long Papers)}, pages 30298--30317, Vienna, Austria. Association for Computational Linguistics.

\bibitem[{Chirkova et~al.(2025)Chirkova, Formal, Nikoulina, and CLINCHANT}]{chirkova2025provence}
Nadezhda Chirkova, Thibault Formal, Vassilina Nikoulina, and St{\'e}phane CLINCHANT. 2025.
\newblock \href {https://openreview.net/forum?id=TDy5Ih78b4} {Provence: efficient and robust context pruning for retrieval-augmented generation}.
\newblock In \emph{The Thirteenth International Conference on Learning Representations}.

\bibitem[{DeepSeek-AI et~al.(2024)DeepSeek-AI, Liu, Feng, Xue, Wang, Wu, Lu, Zhao, Deng, Zhang, Ruan, Dai, Guo, Yang, Chen, Ji, Li, Lin, Dai, Luo, Hao, Chen, Li, Zhang, Bao, Xu, Wang, Zhang, Ding, Xin, Gao, Li, Qu, Cai, Liang, Guo, Ni, Li, Wang, Chen, Chen, Yuan, Qiu, Li, Song, Dong, Hu, Gao, Guan, Huang, Yu, Wang, Zhang, Xu, Xia, Zhao, Wang, Zhang, Li, Wang, Zhang, Zhang, Tang, Li, Tian, Huang, Wang, Zhang, Wang, Zhu, Chen, Du, Chen, Jin, Ge, Zhang, Pan, Wang, Xu, Zhang, Chen, Wu, Ren, Ren, Sha, Fu, Xu, Huang, Zhang, Xie, Zhang, Hao, Gou, Ma, Yan, Shao, Xu, Wu, Zhang, Li, Gu, Zhu, Liu, Li, Xie, Song, Gao, and Pan}]{deepseek_ai_deepseek-v3_2024}
DeepSeek-AI, Aixin Liu, Bei Feng, Bing Xue, Bingxuan Wang, Bochao Wu, Chengda Lu, Chenggang Zhao, Chengqi Deng, Chenyu Zhang, Chong Ruan, Damai Dai, Daya Guo, Dejian Yang, Deli Chen, Dongjie Ji, Erhang Li, Fangyun Lin, Fucong Dai, and 89 others. 2024.
\newblock \href {https://arxiv.org/abs/2412.19437} {Deepseek-v3 technical report}.
\newblock arXiv preprint arXiv:2412.19437.
\newblock CoRR.

\bibitem[{Fu et~al.(2021)Fu, Wang, Zhang, Zhou, and Yan}]{fu-etal-2021-decomposing-complex}
Ruiliu Fu, Han Wang, Xuejun Zhang, Jun Zhou, and Yonghong Yan. 2021.
\newblock \href {https://doi.org/10.18653/v1/2021.findings-emnlp.17} {Decomposing complex questions makes multi-hop {QA} easier and more interpretable}.
\newblock In \emph{Findings of the Association for Computational Linguistics: EMNLP 2021}, pages 169--180, Punta Cana, Dominican Republic. Association for Computational Linguistics.

\bibitem[{Ho et~al.(2020)Ho, Duong~Nguyen, Sugawara, and Aizawa}]{ho-etal-2020-constructing}
Xanh Ho, Anh-Khoa Duong~Nguyen, Saku Sugawara, and Akiko Aizawa. 2020.
\newblock \href {https://doi.org/10.18653/v1/2020.coling-main.580} {Constructing a multi-hop {QA} dataset for comprehensive evaluation of reasoning steps}.
\newblock In \emph{Proceedings of the 28th International Conference on Computational Linguistics}, pages 6609--6625, Barcelona, Spain (Online). International Committee on Computational Linguistics.

\bibitem[{Hurst et~al.(2024)Hurst, Lerer, Goucher, Perelman, Ramesh, Clark, Ostrow, Welihinda, Hayes, Radford, M\k{a}dry, Baker-Whitcomb, Beutel, Borzunov, Carney, Chow, Kirillov, Nichol, Paino, Renzin, Passos et~al.}]{OpenAI_gpt4o_system_card_2024}
Aaron Hurst, Adam Lerer, Adam~P. Goucher, Adam Perelman, Aditya Ramesh, Aidan Clark, AJ~Ostrow, Akila Welihinda, Alan Hayes, Alec Radford, Aleksander M\k{a}dry, Alex Baker-Whitcomb, Alex Beutel, Alex Borzunov, Alex Carney, Alex Chow, Alex Kirillov, Alex Nichol, Alex Paino, and 3 others. 2024.
\newblock \href {https://arxiv.org/abs/2410.21276} {Gpt-4o system card}.
\newblock CoRR.

\bibitem[{Hwang et~al.(2025)Hwang, Cho, Jeong, Song, Han, and Park}]{hwang-etal-2025-exit}
Taeho Hwang, Sukmin Cho, Soyeong Jeong, Hoyun Song, SeungYoon Han, and Jong~C. Park. 2025.
\newblock \href {https://doi.org/10.18653/v1/2025.findings-acl.253} {{EXIT}: Context-aware extractive compression for enhancing retrieval-augmented generation}.
\newblock In \emph{Findings of the Association for Computational Linguistics: ACL 2025}, pages 4895--4924, Vienna, Austria. Association for Computational Linguistics.

\bibitem[{Jeong et~al.(2024)Jeong, Baek, Cho, Hwang, and Park}]{jeong-etal-2024-adaptive}
Soyeong Jeong, Jinheon Baek, Sukmin Cho, Sung~Ju Hwang, and Jong Park. 2024.
\newblock \href {https://doi.org/10.18653/v1/2024.naacl-long.389} {Adaptive-{RAG}: Learning to adapt retrieval-augmented large language models through question complexity}.
\newblock In \emph{Proceedings of the 2024 Conference of the North American Chapter of the Association for Computational Linguistics: Human Language Technologies (Volume 1: Long Papers)}, pages 7036--7050, Mexico City, Mexico. Association for Computational Linguistics.

\bibitem[{Khattab et~al.(2022)Khattab, Santhanam, Li, Hall, Liang, Potts, and Zaharia}]{khattab2022demonstrate}
Omar Khattab, Keshav Santhanam, Xiang~Lisa Li, David Hall, Percy Liang, Christopher Potts, and Matei Zaharia. 2022.
\newblock \href {https://arxiv.org/abs/2212.14024} {Demonstrate-search-predict: Composing retrieval and language models for knowledge-intensive nlp}.
\newblock \emph{arXiv preprint arXiv:2212.14024}.

\bibitem[{Khot et~al.(2023)Khot, Trivedi, Finlayson, Fu, Richardson, Clark, and Sabharwal}]{khot2023decomposed}
Tushar Khot, Harsh Trivedi, Matthew Finlayson, Yao Fu, Kyle Richardson, Peter Clark, and Ashish Sabharwal. 2023.
\newblock \href {https://openreview.net/forum?id=_nGgzQjzaRy} {Decomposed prompting: A modular approach for solving complex tasks}.
\newblock In \emph{The Eleventh International Conference on Learning Representations}.

\bibitem[{Laban et~al.(2024)Laban, Fabbri, Xiong, and Wu}]{laban-etal-2024-summary}
Philippe Laban, Alexander Fabbri, Caiming Xiong, and Chien-Sheng Wu. 2024.
\newblock \href {https://doi.org/10.18653/v1/2024.emnlp-main.552} {Summary of a haystack: A challenge to long-context {LLM}s and {RAG} systems}.
\newblock In \emph{Proceedings of the 2024 Conference on Empirical Methods in Natural Language Processing}, pages 9885--9903, Miami, Florida, USA. Association for Computational Linguistics.

\bibitem[{Lee et~al.(2025{\natexlab{a}})Lee, Jo, Park, and Lee}]{lee-etal-2025-shifting}
Dahyun Lee, Yongrae Jo, Haeju Park, and Moontae Lee. 2025{\natexlab{a}}.
\newblock \href {https://doi.org/10.18653/v1/2025.acl-long.861} {Shifting from ranking to set selection for retrieval augmented generation}.
\newblock In \emph{Proceedings of the 63rd Annual Meeting of the Association for Computational Linguistics (Volume 1: Long Papers)}, pages 17606--17619, Vienna, Austria. Association for Computational Linguistics.

\bibitem[{Lee et~al.(2025{\natexlab{b}})Lee, Oh, Kim, Kim, Park, and Seo}]{lee-etal-2025-rescore}
Dosung Lee, Wonjun Oh, Boyoung Kim, Minyoung Kim, Joonsuk Park, and Paul~Hongsuck Seo. 2025{\natexlab{b}}.
\newblock \href {https://doi.org/10.18653/v1/2025.acl-long.16} {{R}e{SCORE}: Label-free iterative retriever training for multi-hop question answering with relevance-consistency supervision}.
\newblock In \emph{Proceedings of the 63rd Annual Meeting of the Association for Computational Linguistics (Volume 1: Long Papers)}, pages 341--359, Vienna, Austria. Association for Computational Linguistics.

\bibitem[{Liu et~al.(2024)Liu, Lin, Hewitt, Paranjape, Bevilacqua, Petroni, and Liang}]{liu-etal-2024-lost}
Nelson~F. Liu, Kevin Lin, John Hewitt, Ashwin Paranjape, Michele Bevilacqua, Fabio Petroni, and Percy Liang. 2024.
\newblock \href {https://doi.org/10.1162/tacl_a_00638} {Lost in the middle: How language models use long contexts}.
\newblock \emph{Transactions of the Association for Computational Linguistics}, 12:157--173.

\bibitem[{Mei et~al.(2025)Mei, Hu, Fu, Wen, Yang, Wu, Cai, Cai, Gao, Yang, Xie, Shi, Liu, and Qiao}]{mei2025o2searcher}
Jianbiao Mei, Tao Hu, Daocheng Fu, Licheng Wen, Xuemeng Yang, Rong Wu, Pinlong Cai, Xinyu Cai, Xing Gao, Yu~Yang, Chengjun Xie, Botian Shi, Yong Liu, and Yu~Qiao. 2025.
\newblock \href {https://arxiv.org/abs/2505.16582} {O$^2$-searcher: A searching-based agent model for open-domain open-ended question answering}.
\newblock \emph{Preprint}, arXiv:2505.16582.

\bibitem[{Min et~al.(2019)Min, Zhong, Zettlemoyer, and Hajishirzi}]{min-etal-2019-multi}
Sewon Min, Victor Zhong, Luke Zettlemoyer, and Hannaneh Hajishirzi. 2019.
\newblock \href {https://doi.org/10.18653/v1/P19-1613} {Multi-hop reading comprehension through question decomposition and rescoring}.
\newblock In \emph{Proceedings of the 57th Annual Meeting of the Association for Computational Linguistics}, pages 6097--6109, Florence, Italy. Association for Computational Linguistics.

\bibitem[{Pradeep et~al.(2023)Pradeep, Sharifymoghaddam, and Lin}]{pradeep2023rankzephyr}
Ronak Pradeep, Sahel Sharifymoghaddam, and Jimmy Lin. 2023.
\newblock \href {https://arxiv.org/abs/2312.02724} {Rankzephyr: Effective and robust zero-shot listwise reranking is a breeze!}
\newblock \emph{arXiv preprint arXiv:2312.02724}.

\bibitem[{Press et~al.(2023)Press, Zhang, Min, Schmidt, Smith, and Lewis}]{press-etal-2023-measuring}
Ofir Press, Muru Zhang, Sewon Min, Ludwig Schmidt, Noah Smith, and Mike Lewis. 2023.
\newblock \href {https://doi.org/10.18653/v1/2023.findings-emnlp.378} {Measuring and narrowing the compositionality gap in language models}.
\newblock In \emph{Findings of the Association for Computational Linguistics: EMNLP 2023}, pages 5687--5711, Singapore. Association for Computational Linguistics.

\bibitem[{Qi et~al.(2019)Qi, Lin, Mehr, Wang, and Manning}]{qi-etal-2019-answering}
Peng Qi, Xiaowen Lin, Leo Mehr, Zijian Wang, and Christopher~D. Manning. 2019.
\newblock \href {https://doi.org/10.18653/v1/D19-1261} {Answering complex open-domain questions through iterative query generation}.
\newblock In \emph{Proceedings of the 2019 Conference on Empirical Methods in Natural Language Processing and the 9th International Joint Conference on Natural Language Processing (EMNLP-IJCNLP)}, pages 2590--2602, Hong Kong, China. Association for Computational Linguistics.

\bibitem[{Robertson and Zaragoza(2009)}]{robertson2009probabilistic}
Stephen Robertson and Hugo Zaragoza. 2009.
\newblock \href {https://doi.org/10.1561/1500000019} {The probabilistic relevance framework: Bm25 and beyond}.
\newblock \emph{Foundations and Trends in Information Retrieval}, 3(4):333--389.

\bibitem[{Sarthi et~al.(2024)Sarthi, Abdullah, Tuli, Khanna, Goldie, and Manning}]{sarthi2024raptor}
Parth Sarthi, Salman Abdullah, Aditi Tuli, Shubh Khanna, Anna Goldie, and Christopher~D Manning. 2024.
\newblock \href {https://openreview.net/forum?id=GN921JHCRw} {{RAPTOR}: Recursive abstractive processing for tree-organized retrieval}.
\newblock In \emph{The Twelfth International Conference on Learning Representations}.

\bibitem[{Song et~al.(2025)Song, Jiang, Min, Chen, Chen, Zhao, Fang, and Wen}]{R1-searcher}
Huatong Song, Jinhao Jiang, Yingqian Min, Jie Chen, Zhipeng Chen, Wayne~Xin Zhao, Lei Fang, and Ji-Rong Wen. 2025.
\newblock \href {https://github.com/RUCAIBox/R1-searcher} {R1-searcher: Incentivizing the search capability in llms via reinforcement learning}.
\newblock \emph{arXiv preprint arXiv:2503.05592}.

\bibitem[{Sun et~al.(2023)Sun, Yan, Ma, Wang, Ren, Chen, Yin, and Ren}]{sun-etal-2023-chatgpt}
Weiwei Sun, Lingyong Yan, Xinyu Ma, Shuaiqiang Wang, Pengjie Ren, Zhumin Chen, Dawei Yin, and Zhaochun Ren. 2023.
\newblock \href {https://doi.org/10.18653/v1/2023.emnlp-main.923} {Is {C}hat{GPT} good at search? investigating large language models as re-ranking agents}.
\newblock In \emph{Proceedings of the 2023 Conference on Empirical Methods in Natural Language Processing}, pages 14918--14937, Singapore. Association for Computational Linguistics.

\bibitem[{Talmor and Berant(2018)}]{talmor-berant-2018-web}
Alon Talmor and Jonathan Berant. 2018.
\newblock \href {https://doi.org/10.18653/v1/N18-1059} {The web as a knowledge-base for answering complex questions}.
\newblock In \emph{Proceedings of the 2018 Conference of the North {A}merican Chapter of the Association for Computational Linguistics: Human Language Technologies, Volume 1 (Long Papers)}, pages 641--651, New Orleans, Louisiana. Association for Computational Linguistics.

\bibitem[{Tang and Yang(2024)}]{tang2024multihoprag}
Yixuan Tang and Yi~Yang. 2024.
\newblock \href {https://openreview.net/forum?id=t4eB3zYWBK} {Multihop-{RAG}: Benchmarking retrieval-augmented generation for multi-hop queries}.
\newblock In \emph{First Conference on Language Modeling}.

\bibitem[{Team(2025)}]{google_gemini2.5_report_2025}
Google~Gemini Team. 2025.
\newblock \href {https://arxiv.org/abs/2507.06261} {Gemini 2.5: Pushing the frontier with advanced reasoning, multimodality, long context, and next generation agentic capabilities}.
\newblock Technical report, Google / DeepMind.
\newblock CoRR.

\bibitem[{Trivedi et~al.(2022)Trivedi, Balasubramanian, Khot, and Sabharwal}]{trivedi-etal-2022-musique}
Harsh Trivedi, Niranjan Balasubramanian, Tushar Khot, and Ashish Sabharwal. 2022.
\newblock \href {https://doi.org/10.1162/tacl_a_00475} {{M}u{S}i{Q}ue: Multihop questions via single-hop question composition}.
\newblock \emph{Transactions of the Association for Computational Linguistics}, 10:539--554.

\bibitem[{Trivedi et~al.(2023)Trivedi, Balasubramanian, Khot, and Sabharwal}]{trivedi-etal-2023-interleaving}
Harsh Trivedi, Niranjan Balasubramanian, Tushar Khot, and Ashish Sabharwal. 2023.
\newblock \href {https://doi.org/10.18653/v1/2023.acl-long.557} {Interleaving retrieval with chain-of-thought reasoning for knowledge-intensive multi-step questions}.
\newblock In \emph{Proceedings of the 61st Annual Meeting of the Association for Computational Linguistics (Volume 1: Long Papers)}, pages 10014--10037, Toronto, Canada. Association for Computational Linguistics.

\bibitem[{Wang et~al.(2025)Wang, Chen, Yang, Huang, Dou, and Wei}]{wang2025chainofretrieval}
Liang Wang, Haonan Chen, Nan Yang, Xiaolong Huang, Zhicheng Dou, and Furu Wei. 2025.
\newblock \href {https://openreview.net/forum?id=gUPGGCM4WH} {Chain-of-retrieval augmented generation}.
\newblock In \emph{The Thirty-ninth Annual Conference on Neural Information Processing Systems}.

\bibitem[{Wei et~al.(2022)Wei, Wang, Schuurmans, Bosma, Ichter, Xia, Chi, Le, and Zhou}]{wei2022chain}
Jason Wei, Xuezhi Wang, Dale Schuurmans, Maarten Bosma, Brian Ichter, Fei Xia, Ed~H. Chi, Quoc~V. Le, and Denny Zhou. 2022.
\newblock \href {https://openreview.net/pdf?id=_VjQlMeSB_J} {Chain‐of‐thought prompting elicits reasoning in large language models}.
\newblock In \emph{Proceedings of the 36th Conference on Neural Information Processing Systems (NeurIPS 2022)}.

\bibitem[{Weller et~al.(2026)Weller, Boratko, Naim, and Lee}]{weller2026on}
Orion Weller, Michael Boratko, Iftekhar Naim, and Jinhyuk Lee. 2026.
\newblock \href {https://openreview.net/forum?id=k9CzIvzfaA} {On the theoretical limitations of embedding-based retrieval}.
\newblock In \emph{The Fourteenth International Conference on Learning Representations}.

\bibitem[{Xiao et~al.(2023)Xiao, Liu, Zhang, and Muennighoff}]{xiao2023cpack}
Shitao Xiao, Zheng Liu, Peitian Zhang, and Niklas Muennighoff. 2023.
\newblock \href {https://arxiv.org/abs/2309.07597} {C-pack: Packaged resources to advance general chinese embedding}.
\newblock \emph{arXiv preprint arXiv:2309.07597}.

\bibitem[{Xiong et~al.(2021)Xiong, Li, Iyer, Du, Lewis, Wang, Mehdad, Yih, Riedel, Kiela, and Oguz}]{xiong2021answering}
Wenhan Xiong, Xiang Li, Srini Iyer, Jingfei Du, Patrick Lewis, William~Yang Wang, Yashar Mehdad, Scott Yih, Sebastian Riedel, Douwe Kiela, and Barlas Oguz. 2021.
\newblock \href {https://openreview.net/forum?id=EMHoBG0avc1} {Answering complex open-domain questions with multi-hop dense retrieval}.
\newblock In \emph{International Conference on Learning Representations}.

\bibitem[{Xu et~al.(2024)Xu, Shi, and Choi}]{xu2024recomp}
Fangyuan Xu, Weijia Shi, and Eunsol Choi. 2024.
\newblock \href {https://openreview.net/forum?id=mlJLVigNHp} {{RECOMP}: Improving retrieval-augmented {LM}s with context compression and selective augmentation}.
\newblock In \emph{The Twelfth International Conference on Learning Representations}.

\bibitem[{Yang et~al.(2018)Yang, Qi, Zhang, Bengio, Cohen, Salakhutdinov, and Manning}]{yang-etal-2018-hotpotqa}
Zhilin Yang, Peng Qi, Saizheng Zhang, Yoshua Bengio, William Cohen, Ruslan Salakhutdinov, and Christopher~D. Manning. 2018.
\newblock \href {https://doi.org/10.18653/v1/D18-1259} {{H}otpot{QA}: A dataset for diverse, explainable multi-hop question answering}.
\newblock In \emph{Proceedings of the 2018 Conference on Empirical Methods in Natural Language Processing}, pages 2369--2380, Brussels, Belgium. Association for Computational Linguistics.

\bibitem[{Yao et~al.(2023)Yao, Zhao, Yu, Du, Shafran, Narasimhan, and Cao}]{yao2023react}
Shunyu Yao, Jeffrey Zhao, Dian Yu, Nan Du, Izhak Shafran, Karthik~R Narasimhan, and Yuan Cao. 2023.
\newblock \href {https://openreview.net/forum?id=WE_vluYUL-X} {React: Synergizing reasoning and acting in language models}.
\newblock In \emph{The Eleventh International Conference on Learning Representations}.

\bibitem[{Yu et~al.(2024)Yu, Ping, Liu, Wang, You, Zhang, Shoeybi, and Catanzaro}]{yu2024rankrag}
Yue Yu, Wei Ping, Zihan Liu, Boxin Wang, Jiaxuan You, Chao Zhang, Mohammad Shoeybi, and Bryan Catanzaro. 2024.
\newblock \href {https://openreview.net/forum?id=S1fc92uemC} {Rank{RAG}: Unifying context ranking with retrieval-augmented generation in {LLM}s}.
\newblock In \emph{The Thirty-eighth Annual Conference on Neural Information Processing Systems}.

\end{thebibliography}

\clearpage
\appendix

\section{Supplementary Analysis and Results}
\label{app:additional_results_error_analysis}

\subsection{Additional Results}


\paragraph{Passage-Level QA Performance.}
Table~\ref{tab:passage_qa} reports the QA agent’s performance when restricted to answering from the passages retrieved by our framework that contain the supporting facts. Instead of supplying only the supporting facts, we provide entire passages, and the QA agent performs better than full context baseline.

This setting yields stronger QA performance than the full-context baseline, reaching 44.4 EM and 56.81 F1 on 2WikiMultiHopQA and 47.8 EM and 61.15 F1 on HotpotQA. These results highlight that compact yet relevant context is more effective than supplying the entire context, as removing distractors allows the QA agent to better utilize the evidence.

\begin{table}[h]
\centering
\small
\setlength{\tabcolsep}{6pt} 

\begin{tabular}{l|c|c}
\toprule
\textbf{Dataset} & \textbf{EM} & \textbf{F1} \\
\midrule
HotpotQA & 47.8 & {61.15} \\
\midrule
2WikiMultiHopQA & 44.4 & {56.81} \\
\bottomrule
\end{tabular}
\caption{Passage Level QA performance. QA Agent answers the question from the retrieved passages.}
\label{tab:passage_qa}
\end{table}


\paragraph{Fact-Level Retrieval Performance.}
To provide a more fine-grained evaluation of retrieval quality, we also report 
sentence-level (fact-wise) precision, recall, and F1 for our agentic retrieval framework 
on HotpotQA and 2WikiMultiHopQA. Unlike passage-level metrics, which count an entire paragraph  as correct if it contains at least one gold supporting fact, the sentence-level evaluation  requires retrieving the exact gold supporting sentences.

\begin{table}[h]
\centering
\small

\begin{tabular}{l|ccc}
\toprule
\textbf{Dataset} & \textbf{P} & \textbf{R} & \textbf{F1} \\
\midrule
HotpotQA & 56.70 & 75.51 & 64.77 \\
\midrule
2WikiMultiHopQA & 60.81 & 74.42 & 66.93 \\
\bottomrule
\end{tabular}
\caption{Fact-level retrieval performance of our agentic retrieval framework on HotpotQA and 2WikiMultiHopQA. Metrics are precision (P), recall (R), and F1 at the fact (sentence) level.}
\label{tab:sentence_retrieval}
\end{table}

These results show that our framework not only retrieves the correct passages but also identifies the exact supporting facts needed to answer multi-hop questions. This highlights the ability of the Selector$\Leftrightarrow$Adder loop to reduce noise and capture fine-grained evidence, complementing the passage-level metrics reported in the main paper.
\begin{figure*}[ht]
    \centering

    \includegraphics[width=0.9\linewidth]{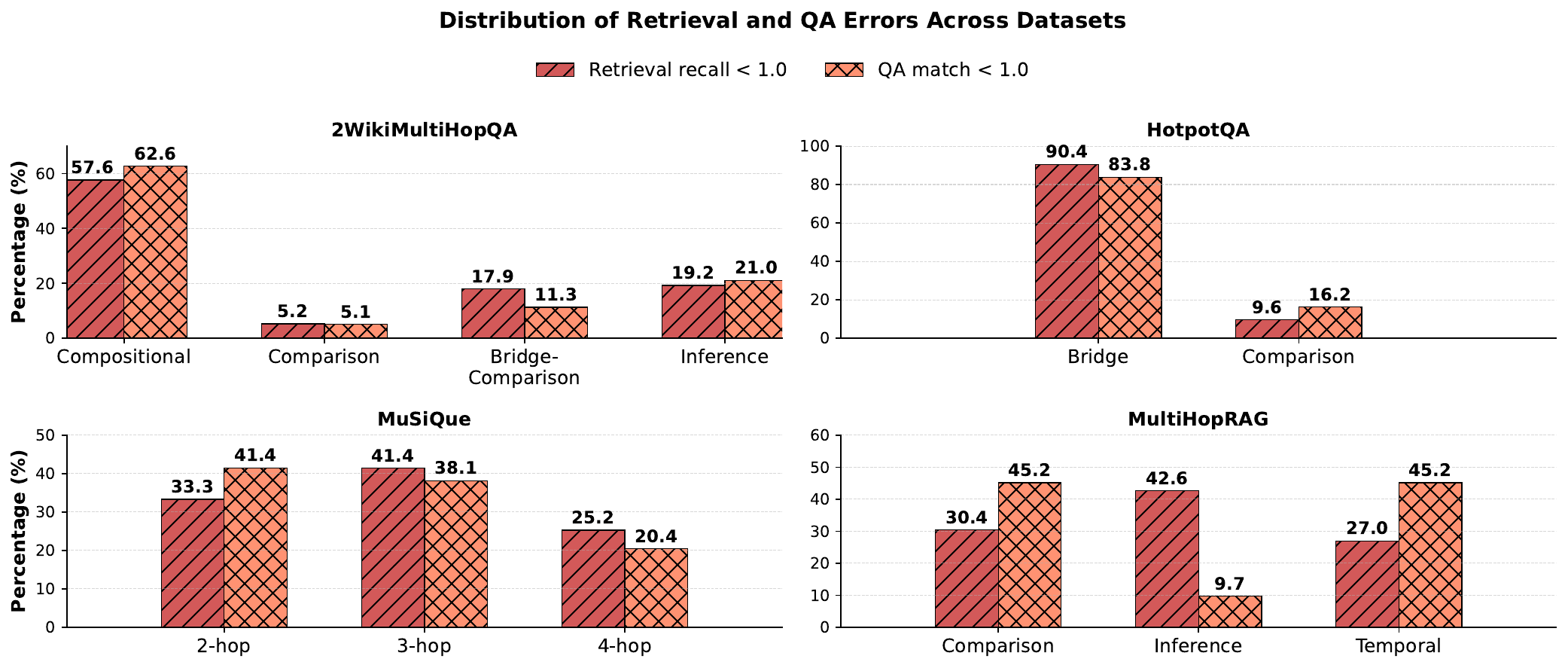}

    \caption{ Error analysis across HotpotQA, 2WikiMultiHopQA, MultiHopRAG and MuSiQue. We compare the distribution of retrieval errors and QA answer errors across question types or hop counts. Across datasets, the error distributions for retrieval and QA are broadly similar, with compositional questions dominating errors in 2WikiMultiHopQA, bridge questions dominating HotpotQA, and 2–3 hop questions accounting for most errors in MuSiQue.}
    \label{fig:error_analysis_all}

\end{figure*}


\subsection{Error Analysis}
\paragraph{Error Analysis (HotPotQA).}
On HotpotQA evaluation set, we examined two error subsets:
(1) items with incomplete evidence retrieval (\textit{recall} $<1.0$) and
(2) items where the QA Agents’s predicted answer did not exactly match the gold answer (\textit{match\_retrieved} $<1.0$).
As shown in Figure~\ref{fig:error_analysis_all}, the majority of errors in both groups are \textbf{bridge} questions
(90.4\% for retrieval failures and 83.8\% for answer mismatches), with the remainder classified as \textbf{comparison}.
Notably, among the items where answer mismatches, about 37\% cases the error occur despite perfect retrieval recall, highlighting that a substantial fraction of failures stem from the QA model rather than the retriever.






\paragraph{Error Analysis (2WikiMultiHopQA).}
In Figure~\ref{fig:error_analysis_all}, we performed a similar analysis on the 2WikiMultiHopQA evaluation set.
Among the questions with incomplete evidence retrieval (\textit{recall}$<1.0$),
the majority are \textbf{compositional} (57.6\%), followed by
\textbf{bridge\_comparison} (17.9\%), \textbf{inference} (19.2\%), and
\textbf{comparison} (5.2\%).
For the questions where the QA Agent’s answer did not exactly match the gold label
(\textit{match\_retrieved}$<1.0$), the distribution is similar:
62.7\% compositional, 11.3\% bridge\_comparison, 21.0\% inference, and 5.1\% comparison. Notably, about 33\% of these answer mismatches occurred despite perfect retrieval recall, indicating that roughly one-third of the answer errors originate from the QA model itself rather than from the retriever.

\paragraph{Error Analysis (MuSiQue).}
On the MuSiQue evaluation set, we analyzed errors by question hop count and summarized it in Figure~\ref{fig:error_analysis_all}. Among the questions with incomplete retrieval (\textit{recall}$<1.0$),
33.3\% require 2 hops, 41.4\% require 3 hops, and 25.3\% require 4 hops.
For the 181 questions where the QA reader’s prediction did not exactly match the gold
(\textit{match\_retrieved}$<1.0$),
41.4\% are 2-hop, 38.1\% are 3-hop, and 20.4\% are 4-hop.
Notably, about 53\% of these answer mismatches occurred despite perfect retrieval recall, revealing that over half of the MuSiQue answer errors stem from the QA model itself rather than from evidence retrieval.






\paragraph{Error Analysis (MultiHopRAG).}
From Figure~\ref{fig:error_analysis_all} we can observe that on the MultiHopRAG evaluation set, among the questions exhibited incomplete retrieval (\textit{recall}$<1.0$) are dominated by
\textbf{inference} queries (42.6\%), with \textbf{comparison} and
\textbf{temporal} queries accounting for 30.4\% and 27.0\%, respectively.
Among the samples where the QA reader’s prediction did not match the gold answer (\textit{match\_retrieved}$<1.0$), the distribution shifts:
\textbf{comparison} and \textbf{temporal} queries each contribute 45.2\%, while only 9.7\% are inference queries.
This indicates that retrieval struggles most with inference-style questions, whereas answer generation errors are more prevalent for comparison and temporal questions even when supporting evidence is retrieved.


\begin{table*}[h]
\centering
\small

\begin{tabular}{l|c|c}
\toprule
\textbf{Dataset} & \textbf{QA Mismatch @ Recall $=1.0$} & \textbf{QA Match @ Recall $<1.0$} \\
\midrule
HotpotQA         & 37.11 & 42.62 \\
\midrule
2WikiMultiHopQA  & 32.68 & 24.45 \\
\midrule
MuSiQue          & 53.59 & 15.15 \\
\midrule
MultiHopRAG      & 12.90 & 64.28 \\
\bottomrule
\end{tabular}
\caption{QA Agent accuracy (\%) under different retrieval coverage.
The first column shows the proportion of questions where the QA agent
did not matched the gold answer when all supporting evidence was retrieved
(\textit{recall}=1.0). The second column shows accuracy where QA Agent matched the gold answer even when evidence recall
was below 1.0.}
\label{tab:qa-vs-recall-all}
\end{table*}

\begin{table*}[h]
\centering
\small

\begin{tabular}{l|c|c|c|c}
\toprule
\textbf{Method} & \textbf{\#LLM Calls} & \textbf{Avg Tokens / Query} & \textbf{Latency (s)} & \textbf{Context Reduction} \\
\midrule
IRCoT  & 8 & 26k & 16 &  - \\
\midrule
CoRAG (L = 10)  & 10 & 21k & 15 & - \\
\midrule
\rowcolor{gray!10}
\textbf{PRISM (Ours)} & 8 & 14k & 8.6 & 73.3\% \\
\bottomrule
\end{tabular}
\caption{Efficiency and scalability comparison. We report number of LLM calls, average tokens per query, inference latency, and reduction in context size on HotPotQA dataset.}
\label{tab:efficiency}
\end{table*}

\paragraph{Cross-Dataset QA Behavior.}
Table~\ref{tab:qa-vs-recall-all} provides a critical observation by comparing QA
performance under two contrasting conditions. The first column (\textit{QA
Mismatch @ Recall $=1.0$}) captures cases where the retriever successfully
collected all gold supporting evidence, yet the QA agent still failed to
produce the correct answer. This highlights reasoning or answer synthesis
errors beyond retrieval. For instance, mismatch rates are high on HotpotQA
(37.1\%) and especially MuSiQue (53.6\%), reflecting the greater reasoning
complexity of these datasets.  

The second column (\textit{QA Match @ Recall $<1.0$}) reports cases where the QA agent produced the correct answer despite missing at least one gold supporting fact. This suggests that some datasets contain redundant evidence, and not all gold-labeled facts are always necessary to answer the question correctly. Notably, MultiHopRAG shows a high proportion of such cases, with 64.3\% still answered correctly, whereas MuSiQue remains much lower at 15.2\%, indicating less redundant evidence and more carefully curated gold supporting facts. 

Overall, these results reveal a nuanced picture: full recall does not
guarantee correctness if reasoning fails, and conversely, partial recall can
still suffice when evidence \textbf{redundancy} exists. This underscores the need for
both high-quality retrieval and robust reasoning in multi-hop QA. 

\paragraph{Analysis on Question Analyzer.}
Due to the multi-hop nature of the questions, we use the Question Analyzer agent to carefully decompose each query into subquestions. In our initial experiments, we observed that
LLMs perform remarkably well at this decomposition step. In addition, the generated subquestions are provided to the Selector and Adder agents as complementary signals along with
the original question, helping them identify and retain the essential evidence needed to answer the query. Since our framework includes a recall-oriented Adder agent for recovery,
errors are less likely to propagate through the iterative loop. To asses robustness, we conducted an error analysis in our initial experiments, where we manually inspected 100 failure
cases across HotpotQA and MuSiQue. Only 8\% stemmed from decomposition errors (e.g., overly granular sub-questions). 

\subsection{Efficiency and Scalability}

Table \ref{tab:efficiency} demonstrate that Our approach is faster, requiring fewer LLM calls (7 vs. 8) and processing less than half the average tokens per query (14k vs. 26k) compared to IRCoT. This leads to a notable reduction in inference latency, achieving 8.6 seconds compared to IRCoT's 16 seconds. Additionally, the method is highly efficient in managing context, resulting in a 73.3\% reduction in context size.

\begin{table}[ht]
\centering
\small
\setlength{\tabcolsep}{4pt}
\begin{tabular}{l|c|c}
\toprule
\textbf{Dataset} & \textbf{IRCoT (\#tok/query)} & \textbf{PRISM (\#tok/query)} \\
\midrule
HotpotQA & 26k & 14k \\
2Wiki & 22.8k & 10.1k \\
MuSiQue & 36k & 28.1k \\
MHRAG & -- & 29.4k \\
\bottomrule
\end{tabular}
\caption{Average token usage per query across datasets. PRISM uses a fixed agentic pipeline with a constant number of LLM calls, resulting in predictable computational cost. Compared to IRCoT, PRISM consistently requires fewer tokens on datasets where both methods are evaluated.}
\label{tab:cost_analysis}
\end{table}

To provide a more comprehensive understanding of computational cost, we extend our analysis beyond HotpotQA to multiple datasets. In PRISM, the overall cost is largely \textbf{independent} of question type or hop count, since all queries are processed using the same fixed agentic pipeline with a constant number of LLM calls. The Question Analyzer decomposes the input into sub-questions without increasing the number of calls, while the main cost driver is the iteration depth of the Selector$\Leftrightarrow$Adder loop. In our default setting, PRISM uses a fixed budget of 8 LLM calls per query, making its computational structure predictable across datasets. Token usage varies primarily with dataset characteristics, especially the size and length of candidate documents. As shown in Table~\ref{tab:cost_analysis}, PRISM consistently requires fewer tokens than strong iterative baselines such as IRCoT while maintaining strong retrieval performance (e.g., 26k vs.\ 14k on HotpotQA, 22.8k vs.\ 10.1k on 2Wiki, and 36k vs.\ 28.1k on MuSiQue). These results highlight that PRISM achieves a favorable efficiency--performance trade-off with lower and more predictable computational cost.

\subsection{Hallucination Analysis}
\label{app:hallucination}

While PRISM primarily targets evidence selection rather than hallucination mitigation, we hypothesize that constructing a compact yet sufficiently complete evidence set can reduce unsupported answer generation by removing distracting passages while preserving the information required for multi-hop reasoning. Since retrieval and QA metrics alone do not directly measure hallucination, we conduct an additional manual analysis to evaluate this effect.

We randomly sample 100 examples from each of HotpotQA and MuSiQue and compare answers generated using the \textbf{Full Evidence} setting with those generated using \textbf{PRISM-retrieved evidence}. We manually label an answer as hallucinated when it contains claims that are not supported by the provided evidence or when the generated answer cannot be justified from the available context. The same evaluation criterion is applied to both settings.

\begin{table}[ht]
\centering
\small
\begin{tabular}{lccc}
\toprule
\textbf{Dataset} & \textbf{Full} & \textbf{PRISM} & \textbf{Reduction} \\
\midrule
HotpotQA & 37\% & 24\% & 13 points \\
MuSiQue  & 48\% & 37\% & 11 points \\
\bottomrule
\end{tabular}
\caption{Manual analysis of unsupported answer generation on 100 randomly sampled examples from each dataset. PRISM-retrieved evidence results in a lower hallucination rate than providing the full retrieved evidence.}
\label{tab:hallucination_analysis}
\end{table}

As shown in Table~\ref{tab:hallucination_analysis}, PRISM substantially reduces unsupported answer generation on both datasets. On HotpotQA, the hallucination rate decreases from 37\% with Full Evidence to 24\% with PRISM, corresponding to a 13-point absolute reduction. On MuSiQue, it decreases from 48\% to 37\%, an 11-point reduction. These results suggest that exposing the QA model to all retrieved passages is not necessarily beneficial: irrelevant or distracting evidence can encourage unsupported reasoning even when the required supporting information is present.

This analysis provides additional evidence that PRISM's iterative precision--recall mechanism improves not only retrieval quality and downstream QA performance, but also the grounding of generated answers. By filtering distractors while recovering potentially missing supporting evidence, PRISM constructs a more focused evidence state from which the QA model can reason. We emphasize, however, that this analysis is based on a manually annotated subset rather than a comprehensive hallucination benchmark, and therefore interpret these results as supporting evidence rather than as a general claim that PRISM eliminates hallucination.


\subsection{Discussion on 2WikiMultiHopQA Performance}
\label{app:2wiki_analysis}

On 2WikiMultiHopQA, PRISM achieves lower downstream QA performance than IRCoT and CoRAG, despite maintaining strong retrieval recall. We believe this gap may arise from both the characteristics of the dataset and differences in the learning and inference settings of the compared methods.

First, 2WikiMultiHopQA often benefits from retaining broader entity-linked evidence during multi-hop reasoning. In such cases, recall-oriented iterative retrieval methods can preserve additional contextual passages that may help establish intermediate entity relations or bridge reasoning steps. PRISM, in contrast, explicitly aims to construct a compact evidence set by filtering potentially distracting passages while recovering evidence judged necessary for the reasoning chain. This design can therefore trade some potentially useful auxiliary context for greater evidence compactness. On datasets such as 2WikiMultiHopQA, where broader contextual coverage can be beneficial, this precision-oriented compression may limit downstream QA performance even when the required supporting evidence is largely retained.

Second, the comparison should be interpreted in light of differences in supervision. PRISM operates in a zero-shot, training-free setting, whereas IRCoT uses few-shot in-context demonstrations and CoRAG relies on task-specific fine-tuning. Consequently, their performance reflects not only differences in retrieval strategy but also additional supervision available during inference or training. In particular, CoRAG can learn dataset-specific retrieval and reasoning behavior through fine-tuning, while PRISM applies the same general inference-time mechanism without parameter updates.


Overall, the 2WikiMultiHopQA results highlight an important trade-off. Recall-heavy or trained approaches can be advantageous when broader contextual retention is particularly useful, whereas PRISM prioritizes a training-free and model-agnostic mechanism for constructing compact yet sufficient evidence. This observation is consistent with PRISM's broader objective of improving evidence efficiency while maintaining competitive downstream QA performance across diverse multi-hop benchmarks.




\section{Extended Literature Review}
\label{rel-work-extended}

\paragraph{Multi-hop QA and Retrieval.} Benchmarks such as HotpotQA \citep{yang-etal-2018-hotpotqa}, 2WikiMultiHopQA \citep{ho-etal-2020-constructing}, and MuSiQue \citep{trivedi-etal-2022-musique} have driven progress in multi-hop reasoning over large corpora. Early retrieval pipelines relied on iterative query rewriting or entity linking (e.g., Graph Retriever \citep{Asai2020Learning}, PathRetriever, MDR \citep{xiong2021answering}), learning to traverse chains of documents. These methods are effective but brittle: errors in early hops often propagate, degrading final performance. \citet{khattab2022demonstrate} introduced the Demonstrate-Search-Predict (DSP) framework, showing how tightly integrating retrieval with reasoning improves performance on knowledge-intensive tasks. \citet{khot2023decomposed} proposed Decomposed Prompting, which breaks complex queries into simpler sub-problems to enhance reasoning. Decomposition has long been used to simplify complex queries, from early heuristic methods \citep{talmor-berant-2018-web, min-etal-2019-multi} to recent LLM-based prompting strategies \citep{ammann2025question, khot2023decomposed, fu-etal-2021-decomposing-complex}. We adopt the decomposition idea in the Question Analyzer agent, which generates sub-questions to guide retrieval, but integrate it within a structured multi-agent loop. While Question Decomposition \citep{ammann2025question, khot2023decomposed} is foundational to multi-hop QA, prior methods typically use decomposition to issue independent search queries, often aggregating noisy results without verification.

Recent work has explored adaptive, iterative, and query-refinement approaches for retrieval-augmented QA. Adaptive-RAG \citep{jeong-etal-2024-adaptive} routes questions to different retrieval strategies based on predicted complexity, while ReSCORE \citep{lee-etal-2025-rescore} improves multi-hop retrieval by training a dense retriever with LLM-derived supervision. RQ-RAG \citep{chan2024rqrag} instead focuses on refining queries through rewriting, decomposition, and disambiguation. In contrast, PRISM is training-free and operates at inference time to refine the evidence set itself through a Selector–Adder loop that explicitly balances precision and recall, yielding compact yet comprehensive evidence for multi-hop QA.

\paragraph{Passage Selection and Reranking.} Parallel work focuses on selecting compact supporting sets. Traditional rerankers score passages independently, which can miss coverage of all reasoning needs. SetR \citep{lee-etal-2025-shifting} addresses this by performing set-wise selection with LLM reasoning, optimizing for coverage across question aspects. Other efforts refine or prune retrieved knowledge before generation, e.g., Provence \citep{chirkova2025provence} and Self-RAG \citep{asai2024selfrag}, both aiming to reduce spurious context. RankZephyr \citep{pradeep2023rankzephyr} introduces an open-source, instruction-tuned model for listwise zero-shot reranking, showing that smaller transparent models can rival or surpass proprietary systems across both in-domain and out-of-domain benchmarks. While RankZephyr focuses on improving listwise reranking efficiency and reproducibility, 
our framework targets multi-hop retrieval with an explicit precision–recall balancing mechanism, addressing a different challenge in evidence selection.  Our Selector agent is conceptually related to set-wise and listwise reranking, but we explicitly decouple precision (Selector) and recall (Adder) in an iterative loop, which provides better control over the precision–recall trade-off in multi-hop QA. RankRAG \citep{yu2024rankrag} unifies context ranking and answer generation by instruction-tuning a single LLM to jointly perform both tasks. While this eliminates the need for a separate reranker, it relies on expensive supervised fine-tuning (SFT) and high-quality training data.

\paragraph{Finetuned Retrievers and Context Optimization}

Recent work has explored optimizing retrieval through context compression, reinforcement learning, and structural alignment. Context Compression methods such as RECOMP \citep{xu2024recomp} and EXIT \citep{hwang-etal-2025-exit} aim to reduce the noise and cost of large context windows by summarizing or compressing retrieved documents before they reach the generation model. While effective for token efficiency, these approaches operate as a post-retrieval step processing whatever the retriever initially returns. Another line of research focuses on Finetuned and RL-based Retrievers. Systems such as CoRAG \citep{wang2025chainofretrieval}, R1-Searcher \citep{R1-searcher}, and O$^2$-Searcher \citep{mei2025o2searcher} utilize supervised fine-tuning (SFT) or reinforcement learning (RL) to train specialized agents that learn to issue search queries or reformulate questions. While these methods achieve strong performance by baking retrieval capabilities into the model weights, they require significant computational resources for training and are often tied to specific model architectures. Our method leverages the inherent reasoning capabilities of off-the-shelf LLMs without requiring task-specific fine-tuning, making it a lightweight, model-agnostic solution that can be easily deployed with any capable instruction-tuned model.

Alignment-Oriented Methods like ARM \citep{chen-etal-2025-retrieve} use a single-shot, solver-based strategy to align questions with structured data layouts. Our framework instead targets multi-hop reasoning over unstructured text, addressing reasoning chains through an iterative precision–recall loop. RAPTOR \citep{sarthi2024raptor} require costly corpus pre-processing and recursive index reconstruction to build tree-organized knowledge structures. In contrast, PRISM achieves its strong performance purely through \textit{inference-time} agentic interaction over standard flat indices, avoiding this significant upfront computational overhead.

\paragraph{LLMs, Chain-of-Thought, and Agentic Retrieval.} The rise of large language models enabled retrieval-augmented reasoning via prompting. Frameworks such as ReAct \citep{yao2023react} and Self-Ask \citep{press-etal-2023-measuring} interleave reasoning with tool calls, letting LLMs pose sub-questions and fetch evidence on the fly. More recently, IRCoT \citep{trivedi-etal-2023-interleaving} explicitly couples CoT reasoning with iterative retrieval, substantially improving recall by accumulating passages across reasoning steps. However, IRCoT emphasizes recall over precision, often yielding large evidence sets with many distractors. Our approach synthesizes three lines of research—LLM-guided retrieval, set-wise selection, and decomposition—into a unified framework. Unlike prior systems, we combine explicit decomposition with an iterative precision-and-recall retrieval loop, achieving high recall while maintaining a compact and noise-resistant evidence set.

\onecolumn




\definecolor{promptbg}{RGB}{250,251,253}
\definecolor{promptframe}{RGB}{170,180,195}
\definecolor{prompttitle}{RGB}{40,60,90}
\definecolor{prompttext}{RGB}{30,30,30}
\definecolor{promptaccent}{RGB}{70,100,150}


\section{Illustrative Example}
\label{app:illustrative_example}
To provide intuition on how our retrieval framework operates, we present an example from the 2WikiMultiHopQA dataset.  
This case highlights the iterative interaction between the Adder and Selector agents and the resulting evidence refinement.

\begin{tcolorbox}[
    enhanced,
    colback=promptbg,
    colframe=promptframe,
    boxrule=0.5pt,
    arc=2pt,
    left=8pt,
    right=8pt,
    top=6pt,
    bottom=6pt,
    title=\textbf{Illustrative Example (ID: e95acdbc085f11ebbd5dac1f6bf848b6)},
    coltitle=black,
    fonttitle=\bfseries\large,
    fontupper=\footnotesize\ttfamily\color{prompttext},
    breakable
]
\textcolor{promptaccent}{\textbf{Question:}}  
Which film has the director who died earlier, \textit{Deuce High} or \textit{The King Is The Best Mayor}?

\textcolor{promptaccent}{\textbf{Gold Answer:}}  
\textit{The King Is The Best Mayor}

\textcolor{promptaccent}{\textbf{Question Analyzer Output:}}
\texttt{
["Who directed \textit{Deuce High}?", "Who directed \textit{The King Is The Best Mayor}?",
"When did the director of \textit{Deuce High} die?", "When did the director of \textit{The King Is The Best Mayor} die?", "Which director died earlier?"]
}

\textcolor{promptaccent}{\textbf{Initial Evidence (Zero-shot LLM):}}  

\verb|[["Deuce High", 0], ["The King is the Best Mayor", 0]]|

\textcolor{promptaccent}{\textbf{Iteration 1:}}  

\textbf{Adder:}  
\verb|[["Deuce High", 0], ["The King is the Best Mayor", 0], ["Richard Thorpe", 0], ["Rafael Gil", 0]]|

\textbf{Selector:}  
Same as Adder.

\textcolor{promptaccent}{\textbf{Iteration 2:}}  

\textbf{Adder:}  
Added \verb|["Richard Thorpe", 1]| and \verb|["Rafael Gil", 1]|.

\textbf{Selector:}  
Pruned to -->  
\texttt{[
["Deuce High", 0], ["The King is the Best Mayor", 0],
["Richard Thorpe", 0], ["Rafael Gil", 0]
]}

\textcolor{promptaccent}{\textbf{Iteration 3:}}  

\textbf{Adder:}  
Re-added --> \verb|["Richard Thorpe", 1]| and \verb|["Rafael Gil", 1]|.

\textbf{Selector:}  
Retained the compact evidence set.

\textcolor{promptaccent}{\textbf{Final Retrieved Evidence:}}  

\verb|[["Deuce High", 0], ["The King is the Best Mayor", 0], ["Richard Thorpe", 0], ["Rafael Gil", 0]]|

\textcolor{promptaccent}{\textbf{Gold Supporting Facts:}}  
Same as the final retrieved evidence.

\textcolor{promptaccent}{\textbf{Retrieval Metrics:}}  

Precision = 1.0, Recall = 1.0, F1 = 1.0, False Positive Rate = 0.0
\end{tcolorbox}

This example illustrates how the Selector--Adder loop converges to the correct supporting evidence set. Although downstream QA still requires subtle reasoning over death dates, the retrieval framework achieves perfect precision and recall, demonstrating its ability to isolate the necessary facts with minimal noise.


\newpage
\section{Prompts}
\label{app:prompts}

\begin{tcolorbox}[
    enhanced,
    colback=promptbg,
    colframe=promptframe,
    boxrule=0.6pt,
    arc=2pt,
    left=8pt,
    right=8pt,
    top=6pt,
    bottom=6pt,
    title=\textbf{Question Analyzer Prompt Template},
    coltitle=black,
    fonttitle=\bfseries\normalsize,
    fontupper=\footnotesize\ttfamily\color{prompttext},
    breakable
]
You are a \textbf{Question Analyzer Agent} in a multi-hop question answering system.

Your task is to analyze a complex multi-hop question and decompose it into a sequence of concise and meaningful subquestions that reflect the reasoning steps required to answer the original question.

Extract subquestions by identifying:
\begin{itemize}
    \item key entities (e.g., persons, organizations, locations, dates),
    \item important nouns or noun phrases (e.g., titles, concepts, objects),
    \item logical or temporal relationships (e.g., comparison, causality, sequence), and
    \item specific conditions or constraints in the question.
\end{itemize}

Each subquestion should retrieve or verify one specific piece of evidence that contributes to the final answer. Return an ordered list of subquestions that forms a clear reasoning path from the question to the answer. Keep each subquestion short, specific, and unambiguous.

\textcolor{promptaccent}{\textbf{Example Input:}}

\textit{Which actor played the brother of the character who was portrayed by the same actress that starred in Legally Blonde?}

\textcolor{promptaccent}{\textbf{Example Output:}}
\begin{enumerate}
    \item Who starred in \textit{Legally Blonde}?
    \item What character did that actress portray in another film?
    \item Who played the brother of that character?
\end{enumerate}

\textcolor{promptaccent}{\textbf{Output Format:}}

Return a JSON object in the following format:

\verb|{"Subquestions": ["...", "..."]}|

\textcolor{promptaccent}{\textbf{Question:}} \verb|[QUESTION]|
\end{tcolorbox}


\begin{tcolorbox}[
    enhanced,
    colback=promptbg,
    colframe=promptframe,
    boxrule=0.5pt,
    arc=2pt,
    left=8pt,
    right=8pt,
    top=6pt,
    bottom=6pt,
    title=\textbf{Selector Agent Prompt Template},
    coltitle=black,
    fonttitle=\bfseries\large,
    fontupper=\footnotesize\ttfamily\color{prompttext},
    breakable
]
You are a \textbf{Selector Agent} in a multi-hop question answering system.

Your goal is to \textbf{maximize precision and minimize false positives}.

You are given:
\begin{itemize}
    \item a complex multi-hop question,
    \item a set of subquestions representing the reasoning steps,
    \item a list of candidate facts, each represented as \verb|[title, sentence_index]|, and
    \item a set of \textbf{currently selected evidence} sentences.
\end{itemize}

Your task is to carefully remove \textbf{only those evidence items that are definitely irrelevant} for answering any of the subquestions.

\textcolor{promptaccent}{\textbf{Important Guidelines:}}
\begin{itemize}
    \item Do \textit{not} remove sentences that are partially relevant or may help bridge reasoning steps.
    \item Retain sentences containing named entities, dates, or events referenced in the question.
    \item Do not add new items or regenerate content. Work strictly with the given evidence list.
\end{itemize}

Return only the pruned list of \verb|[title, sentence_index]| pairs, with no explanation. The sentence index starts from 0 for each paragraph.

\textcolor{promptaccent}{\textbf{Question:}} \verb|[QUESTION]|

\textcolor{promptaccent}{\textbf{Subquestions:}} \verb|[SUBQUESTION]|

\textcolor{promptaccent}{\textbf{Candidates:}} \verb|[CANDIDATES]|

\textcolor{promptaccent}{\textbf{Current Selected Evidence:}} \verb|[CURRENT_EVIDENCE]|

\textcolor{promptaccent}{\textbf{Output:}} Return only the updated list with clearly irrelevant sentences removed.
\end{tcolorbox}

\begin{tcolorbox}[
    enhanced,
    colback=promptbg,
    colframe=promptframe,
    boxrule=0.5pt,
    arc=2pt,
    left=8pt,
    right=8pt,
    top=6pt,
    bottom=6pt,
    title=\textbf{Adder Agent Prompt Template},
    coltitle=black,
    fonttitle=\bfseries\large,
    fontupper=\footnotesize\ttfamily\color{prompttext},
    breakable
]
You are an \textbf{Adder Agent} in a multi-hop question answering system.

Your goal is to \textbf{maximize recall while minimizing false positives}.

You are given:
\begin{itemize}
    \item a complex multi-hop question,
    \item a set of subquestions representing the reasoning steps,
    \item a list of candidate passages and sentences, and
    \item a set of \textbf{currently selected evidence} items, each represented as \verb|[title, sentence_index]|.
\end{itemize}

\textcolor{promptaccent}{\textbf{Your Task:}}

Add only those candidate sentences that are likely to support answering any of the subquestions.

\textcolor{promptaccent}{\textbf{Do Not Add:}}
\begin{itemize}
    \item vague, unrelated, or overly general sentences,
    \item off-topic facts or irrelevant entities, and
    \item duplicates or sentences that overlap substantially with the existing evidence.
\end{itemize}

\textcolor{promptaccent}{\textbf{Focus On:}}
\begin{itemize}
    \item bridging facts that connect entities across subquestions,
    \item sentences containing named entities, dates, definitions, or relationships mentioned in the question, and
    \item factual statements that clearly contribute to the reasoning chain.
\end{itemize}

Do not remove or modify the currently selected evidence. Return a combined list of both current and newly added items as \verb|[title, sentence_index]| pairs, with no explanation. The sentence index starts from 0 for each paragraph.

\textcolor{promptaccent}{\textbf{Question:}} \verb|[QUESTION]|

\textcolor{promptaccent}{\textbf{Subquestions:}} \verb|[SUBQUESTION]|

\textcolor{promptaccent}{\textbf{Candidates:}} \verb|[CANDIDATES]|

\textcolor{promptaccent}{\textbf{Current Selected Evidence:}} \verb|[CURRENT_EVIDENCE]|

\textcolor{promptaccent}{\textbf{Output:}} Return the updated evidence list with only relevant additions.
\end{tcolorbox}


\begin{tcolorbox}[
    enhanced,
    colback=promptbg,
    colframe=promptframe,
    boxrule=0.5pt,
    arc=2pt,
    left=8pt,
    right=8pt,
    top=6pt,
    bottom=6pt,
    title=\textbf{Answer Generator Prompt Template},
    coltitle=black,
    fonttitle=\bfseries\large,
    fontupper=\footnotesize\ttfamily\color{prompttext},
    breakable
]
You are a \textbf{question answering agent}.

Given a question and the supporting evidence, provide a concise, factual, and short answer based \textbf{only} on the evidence. Do not include any additional words or explanation. If the answer cannot be determined from the evidence, reply with \verb|'Not Answerable'|.

\textcolor{promptaccent}{\textbf{Question:}} \verb|[QUESTION]|

\textcolor{promptaccent}{\textbf{Evidence:}} \verb|[EVIDENCE]|

\textcolor{promptaccent}{\textbf{Answer:}}
\end{tcolorbox}


\section{LLM Usage}

Large language models were used only to improve the clarity and readability of this manuscript. After we completed the technical content, an LLM suggested minor edits for grammar, style, and phrasing, which we reviewed before inclusion. All conceptual contributions, analyses, and implementations are entirely our own. LLMs were also occasionally used to assist with debugging code, and all program logic was verified and integrated by the authors.

\end{document}